\documentclass{article} 
\usepackage{iclr2027_conference,times}

\usepackage{amsmath,amsfonts,bm}

\def\eqref#1{equation~\ref{#1}}

\def\1{\bm{1}}

\DeclareMathAlphabet{\mathsfit}{\encodingdefault}{\sfdefault}{m}{sl}
\SetMathAlphabet{\mathsfit}{bold}{\encodingdefault}{\sfdefault}{bx}{n}

\usepackage{hyperref}
\usepackage{url}

\usepackage{graphicx}
\usepackage{multirow}
\usepackage{booktabs}
\usepackage{adjustbox}
\usepackage{algorithm}
\usepackage{algpseudocode}
\usepackage{caption}
\usepackage{enumitem}
\usepackage{placeins}
\usepackage{amsthm}
\usepackage{wrapfig}
\usepackage{array}
\usepackage{xcolor}

\theoremstyle{plain}
\newtheorem{proposition}{Proposition}
\theoremstyle{definition}
\newtheorem{definition}{Definition}

\newcommand{\runemazes}{Rune-Mazes}
\newcommand{\cInv}{\texttt{Invert}}
\newcommand{\cNoInv}{\texttt{NoOp + Invert}}
\newcommand{\cSkInv}{\texttt{SkipNext + Invert}}
\newcommand{\cRes}{\texttt{3 Invert + ResetRules}}
\newcommand{\cRSI}{\texttt{RepeatPrev + SkipNext + Invert}}

\newcommand{\aleralgobox}[3]{%
  \noindent
  \setlength{\fboxsep}{4pt}%
  \fbox{%
    \begin{minipage}[t]{\dimexpr0.96\linewidth-2\fboxsep-2\fboxrule\relax}
      \captionsetup{type=algorithm, hypcap=false, font=small, labelfont=bf, skip=2pt}
      \vspace{1pt}
      \captionof{algorithm}{#1}
      \label{#2}
      \vspace{2pt}
      \hrule
      \vspace{4pt}
      \footnotesize
      #3
      \vspace{1pt}
    \end{minipage}%
  }%
}

\definecolor{cites}{RGB}{204,0,102}

\hypersetup{
    colorlinks=true,
    citecolor=cites,
    linkcolor=black,
    urlcolor=cites
}

\title{ALER: Adaptive Learnable Experience \\ Rewriting for Reinforcement Learning}

\author{
  Oleg Shchendrigin$^{1,2}$ \quad Egor Cherepanov$^{2,3}$ \quad Aleksandr I. Panov$^{2,3}$ \quad Alexey K. Kovalev$^{2,3}$\\
  $^{1}$Innopolis University, Innopolis, Russia \quad $^{2}$MIRIAI, Moscow, Russia \\
  $^{3}$Cognitive AI Systems Lab, Moscow, Russia \\
  \texttt{shchendrigin.o@miriai.org}
}

\iclrfinalcopy 
\begin{document}
\raggedbottom

\maketitle

\begin{abstract}
In partially observable reinforcement learning (RL), a later observation can make stored information obsolete or change what it implies for the next decision. Memory architectures and benchmarks for RL mostly test retention, the ability to keep information unchanged until it is needed. We formalize two further requirements. Rewriting sets the decision-relevant content to a value independent of the old one, and experience fusion transforms the old content by a rule that a later observation specifies. For tasks built from such updates, we count the memory states that a solution needs, and several baselines reach their lowest success rates on compositions that need more states. We introduce ALER (Adaptive Learnable Experience Rewriting), an agent that pairs an LSTM with a slot memory. An independently addressed Gumbel-Softmax write that concentrates its weight on one slot overwrites that slot, and a learned gate fuses the retrieved content with the recurrent state before the policy and value heads. We also introduce \runemazes{}, three environments in which rune observations invert, cancel, reset, or repeat updates of a hidden cue under vector and pixel observations. Against seven baselines, ALER reaches a success rate of at least $0.82$ in all sixteen Endless T-Maze configurations and at least $0.99$ on all five Rune T-Maze compositions, and it has the highest mean success rate on four-branch Rune Multi-Corridor with an \cInv{} rune. On pixel-based Rune MiniGrid Memory, it has a higher mean success rate than PPO-LSTM in eight of ten configurations. 
\textbf{Project page: \url{https://quartz-admirer.github.io/ALER-Adaptive-Learnable-Experience-Rewriting/}}.
\end{abstract}

\section{Introduction}
\begin{wrapfigure}{r}{0.40\textwidth}
  \centering
  \vspace{-5em}
  \includegraphics[width=0.40\textwidth]{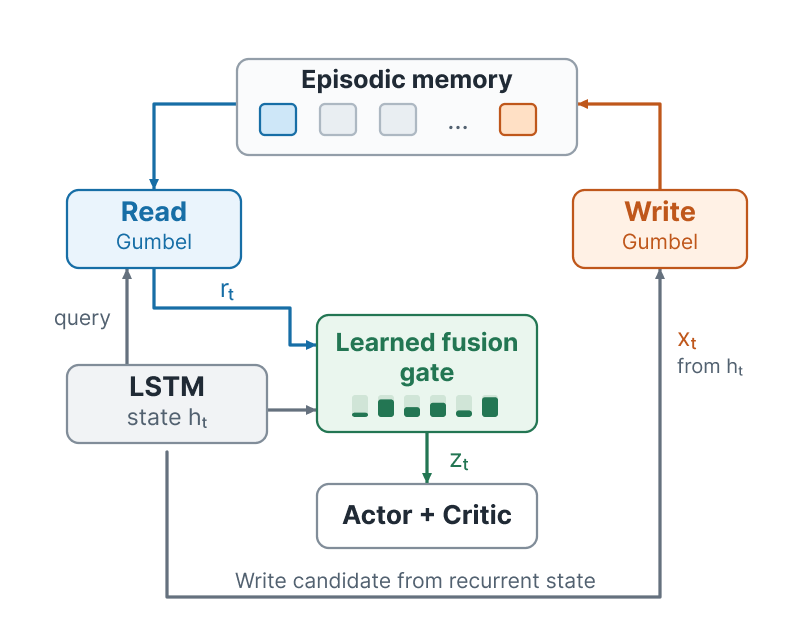}
  \vspace{-2em}
  \caption{\textbf{ALER overview.} The recurrent state $h_t$ queries a slot memory, a learned gate fuses the retrieved vector $r_t$ with $h_t$ into $z_t$ for the actor and critic, and an independently addressed write stores the write candidate $x_t$.}
  \label{fig:visual-abstract}
\vspace{-2em}
\end{wrapfigure}

An agent in a partially observable environment has to keep information that its current observation no longer shows~\citep{kaelbling1998planning}. Consider a T-Maze in which the agent sees a cue at the start of a corridor and must turn toward the side it indicates at the junction. If a later corridor shows a new cue, the old cue becomes obsolete and the agent has to rewrite it. If the corridor contains a rune that inverts the cue, the agent has to turn the other way, which it can do only if it still holds the cue and applies the inversion. The same stored cue thus has to be kept, rewritten, or transformed, depending on what the agent observes later.

Memory architectures for RL are mostly built and evaluated for the first of these cases. Recurrent agents~\citep{ni2022recurrentstrong,cayci2024recurrentnatural} compress the history into one evolving hidden state, and state-space models~\citep{lu2023structuredstatespace} extend the effective horizon with structured recurrent dynamics. Memory-augmented transformers such as GTrXL~\citep{parisotto2020stabilizing}, HCAM~\citep{lampinen2021hcam}, and DTQN~\citep{esslinger2022dtqn} attend over cached activations of the past, and their gains in RL can come from better memory~\citep{ni2023whendotransformers}. Benchmarks for these agents emphasize retention over long horizons~\citep{wang2025investigatingmemoryrlpopgym,pleines2025memorygymendlesstasks}. Endless T-Maze and Color-Cubes~\citep{shchendrigin2026memory} add repeated rewriting of stored cues. To our knowledge, no benchmark isolates the transformation of stored content by later observations.

We make the three requirements precise in \autoref{sec:theory}. An observation calls for retention, rewriting, or experience fusion when it maps the decision-relevant content to itself, to a value independent of it, or to a different function of it. For tasks composed of such updates, we count the memory states that a solution needs. The count separates two solutions. An online solution folds every rune into the stored cue, and a deferred solution keeps the cue unchanged and tracks the runes apart from it. Some compositions force every solution to keep the original cue next to the transformed one.

Adaptive Learnable Experience Rewriting (ALER, \autoref{fig:visual-abstract}) is built around this division of work. An LSTM tracks quantities that change at every step. A slot keeps its content while later writes put little total weight on it, and an independently addressed write that concentrates its weight on one slot overwrites that slot in one step. A learned gate combines the retrieved content with the recurrent state.

To test the three requirements, we introduce \runemazes{}, in which runes invert, cancel, reset, or repeat updates of a cue in vector corridors (Rune T-Maze and Rune Multi-Corridor) and in a pixel-based MiniGrid maze (Rune MiniGrid Memory). Against seven baselines, ALER reaches a success rate of at least $0.82$ in every Endless T-Maze configuration and at least $0.99$ on every Rune T-Maze composition. It keeps its success up to three times the training corridor length and leads PPO-LSTM in eight of ten Rune MiniGrid Memory configurations. Probes show that ALER stores a new cue with one write and that its policy uses the memory read.

Our contributions are the following.
\begin{itemize}[topsep=2pt,itemsep=1pt,parsep=0pt,leftmargin=1.5em]
    \item A formalization of retention, rewriting, and experience fusion as classes of target updates, with bounds on the memory that a cue-decision task needs (\autoref{sec:theory}).
    \item ALER, which combines an LSTM with an independently addressed slot memory and a learned fusion gate, with a characterization of its write operation (\autoref{sec:aler}).
    \item \runemazes{}, three environments whose rune compositions have known memory requirements, under vector and pixel observations (\autoref{sec:environments}).
    \item A comparison with seven baselines and an analysis of what ALER stores and how its policy uses it (Sections~\ref{sec:results} and~\ref{sec:mechanism}).
\end{itemize}

\section{Related Work}
\label{sec:related}
\paragraph{Recurrent and sequence memory.}
LSTM policies~\citep{hochreiter1997lstm,heess2015memory}, including PPO-LSTM~\citep{schulman2017ppo}, remain strong POMDP baselines~\citep{ni2022recurrentstrong,cayci2024recurrentnatural}. Recurrent trace units train a recurrent state online by exact real-time recurrent learning in streaming RL~\citep{farr2026streaming}. Transformer-XL~\citep{dai2019transformerxlattentivelanguagemodels}, GTrXL~\citep{parisotto2020stabilizing}, HCAM~\citep{lampinen2021hcam}, Decision Transformer~\citep{chen2021decisiontransformer}, and RATE~\citep{cherepanov2023rate} attend over a context cache, and S5~\citep{smith2023simplified,lu2023structuredstatespace} uses structured state-space recurrence. In these models, stored content changes inside the recurrent update or the attention computation. ALER keeps a recurrent state and adds a slot memory whose content changes only through explicit writes.

\paragraph{External memory and write rules.}
Neural Turing Machines~\citep{graves2014neuralturingmachines} and DNCs~\citep{graves2016dnc} write to differentiable memory with erase and add operations, and RL-NTM~\citep{zaremba2015rlntm} trains discrete memory access with reinforcement signals. MERLIN~\citep{wayne2018unsupervisedpredictivememorygoaldirected}, working and episodic memory agents~\citep{fortunato2020generalizationreinforcementlearnersworking}, Neural Episodic Control~\citep{pritzel2017neuralepisodiccontrol}, and Hopfield-style episodic memory~\citep{NEURIPS2024_b528459c} bring external memory into RL. Fast and Forgetful Memory (FFM)~\citep{morad2023reinforcement} and Stable Hadamard Memory (SHM)~\citep{le2025stable} apply structured decay and calibration updates to a matrix memory. TARDIS~\citep{gulcehre2017wormhole} also uses near-discrete Gumbel-Softmax addressing, ties its write location to the previously read slot, and is trained with supervised sequence objectives. ELMUR~\citep{cherepanov2025elmur} updates layer-local transformer memory offline by imitation for retention tasks. Memo~\citep{gupta2025memo} trains RL agents to write summary tokens, and LLM agents learn memory operations from data~\citep{zhang2025learn} or receive reminders from a memory agent~\citep{wu2026remember}. ALER learns separate read and write distributions inside an online PPO agent, so one step can read one slot and overwrite another, and a learned gate controls how the read enters the policy.

\paragraph{Memory benchmarks.}
Partially observable versions of MuJoCo~\citep{todorov2012mujoco} and DeepMind Control~\citep{tassa2018deepmindcontrolsuite,ni2022recurrentstrong}, POPGym~\citep{morad2023popgymbenchmarkingpartiallyobservable}, Atari~\citep{bellemare2013ale}, Obstacle Tower~\citep{juliani2019obstacletower}, Procgen~\citep{cobbe2020procgen}, POPGym Arcade~\citep{wang2025investigatingmemoryrlpopgym}, DeepMind Lab~\citep{beattie2016deepmindlab}, Memory Maze~\citep{pasukonis2022memorymaze}, MiniGrid~\citep{chevalierboisvert2023minigridminiworldmodular}, Memory Gym~\citep{pleines2025memorygymendlesstasks}, MIKASA~\citep{cherepanov2025mikasa}, and image-goal navigation~\citep{mezghani2022memory} test memory in RL. Most of their tasks test retention, and some, such as Count Recall in POPGym, require continual updates. Endless T-Maze and Color-Cubes~\citep{shchendrigin2026memory} test the rewriting of stored cues. WorldMemArena~\citep{liu2026worldmemarena} evaluates the writing, maintenance, retrieval, and use of multimodal agent memory. \runemazes{} add transformations of a stored cue by later runes, and each of its tasks comes with a count of the memory states that a solution needs.

\section{Memory Update Requirements}
\label{sec:theory}
This section assigns every observation in our environments to an update class and counts how much memory a task needs. \autoref{sec:results} relates these counts to the success of the baselines, and \autoref{sec:mechanism} compares ALER's representation with the two solutions that they distinguish.

\paragraph{Setup.}
We consider an episodic POMDP $(\mathcal{S},\mathcal{A},\mathcal{O},P,R,\gamma)$ in which the policy $\pi(a_t\mid h_t)$ conditions on the history $h_t=(o_{1:t},a_{1:t-1})$. In a \emph{cue-decision task}, the agent receives reward $+1$ at a decision step $\tau$ (the junction) if its action matches a target $y_\tau\in\mathcal{Y}$ with $|\mathcal{Y}|=K$, and $-1$ otherwise. For $t<\tau$, the target $y_t$ is the value that would be correct if the decision followed $o_t$. It starts at a cue $c=y_1$ shown in the first observation. We assume the following.

\textbf{(D)} The target $y_\tau$ is uniform on $\mathcal{Y}$ and independent of the decision-step observation $o_\tau$.

In all our environments, the cue is uniform, independent of the rune placement, and masked after the first step, and the junction observation does not depend on it, so (D) holds.

\paragraph{Update classes.}
Each observation updates the target, $y_t=F_{o_t}(y_{t-1})$, where the map $F_{o_t}\colon\mathcal{Y}\to\mathcal{Y}$ may depend on the cue $c$ and on a \emph{rule state} $\rho_{t-1}$. The rule state holds the task variables that determine how later observations act, such as a pending \texttt{SkipNext}, and evolves with the observations alone. In Rune T-Maze, the rule state is the skip flag and the last effective rune, and the invert flag is part of the target, $y=c\oplus\mathit{invert}$.

\begin{definition}[Update requirements]
\label{def:classes}
Given $c$ and $\rho_{t-1}$, the observation $o_t$ requires \emph{retention} if $F_{o_t}$ is the identity on $\mathcal{Y}$, \emph{rewriting} if $F_{o_t}$ is constant, and \emph{experience fusion} otherwise.
\end{definition}

Rewriting lets an agent discard the old target. Fusion requires the agent to keep the old target and change it. \autoref{tab:update-classes} (left) classifies the observations of our environments. \texttt{SkipNext} cancels the next rune, \texttt{RepeatPrev} acts as the last effective rune, and \texttt{ResetRules} rewrites the target with the initial cue, so solving a composition that contains it requires retaining the cue.

\begin{table}[t]
\centering
\small
\caption{\textbf{Update effects and memory counts of the \runemazes{} tasks.} Left: effect of each observation on the target $y$ and the rule state. Right: for each rune composition, the number of rune states $Q$ and the number of target states $Q^\star$ over all rune sequences (\autoref{prop:composition}), and the exact counts $Q_{\mathrm{ep}}$ and $Q^\star_{\mathrm{ep}}$ for sampled episodes (\autoref{app:proofs}). The deferred solution uses $KQ$ states. Rune Multi-Corridor has $K$ branches.}
\vspace{-1em}
\label{tab:update-classes}
\footnotesize
\begin{minipage}[t]{0.50\linewidth}
\centering
\setlength{\tabcolsep}{3pt}
\begin{tabular}{@{}lll@{}}
\toprule
\textbf{Observation} & \textbf{Effect} &  \\
\midrule
Corridor step, \texttt{NoOp} & none \\
New cue (Endless) & $y\leftarrow c'$ \\
\texttt{Invert} & $y\leftarrow K{-}1{-}y$\\
\texttt{SkipNext} & cancels next rune \\
\texttt{ResetRules} & $y\leftarrow c$, flags cleared \\
\texttt{RepeatPrev} & repeats last rune \\
\bottomrule
\end{tabular}
\end{minipage}\hfill
\begin{minipage}[t]{0.49\linewidth}
\centering
\setlength{\tabcolsep}{3pt}
\resizebox{\linewidth}{!}{%
\begin{tabular}{@{}lcccc@{}}
\toprule
\textbf{Composition} & $Q$ & $Q^\star$ & $Q_{\mathrm{ep}}$ & $Q^\star_{\mathrm{ep}}$ \\
\midrule
\cInv & 2 & 2 & 2 & 2 \\
\cNoInv & 2 & 2 & 2 & 2 \\
\cSkInv & 4 & 4 & 3 & 4 \\
\cRes & 2 & 4 & 2 & 4 \\
\cRSI & 10 & 10 & 4 & 6 \\
\midrule
Multi-Corridor, \cInv & 2 & $K$ & 2 & $K$ \\
\bottomrule
\end{tabular}}
\end{minipage}
\vspace{-1em}
\end{table}

\paragraph{Memory for one decision.}
The first result bounds the success of any agent by the number of values its memory can take at the junction.

\begin{proposition}[Memory for a decision]
\label{prop:bound}
Under (D), let the action at $\tau$ depend only on $o_\tau$ and a memory variable $m_\tau$ with $M$ possible values. Then a correct decision has probability at most $\min(1,M/K)$.
\end{proposition}

Each memory value can point to at most one target, and (D) makes every target equally likely. A memoryless policy therefore succeeds with probability at most $1/K$, the \textbf{Random} column of \autoref{tab:rune_multi_results}, and a reliable choice among $K$ branches needs at least $\log_2 K$ bits of memory (\autoref{app:proofs}).

\paragraph{Memory for a composed update.}
A rune composition determines how much memory the target itself requires. For a composition with rune alphabet $\Sigma$, let $Q(\Sigma)$ be the number of states of the minimal \emph{rune automaton}, which reads runes from $\Sigma$ and outputs whether the target is inverted relative to the cue, and call its state the \emph{rune state}. Let $Q^\star(\Sigma)$ be the number of states that the minimal automaton reading the cue and then runes from $\Sigma$ occupies after the cue, with the target as output. Both counts describe the update rule over all rune sequences in $\Sigma^\ast$. Sampled episodes show each rune candidate at most once, and for them we compute the exact counts $Q_{\mathrm{ep}}\le Q$ and $Q^\star_{\mathrm{ep}}\le Q^\star$ (\autoref{app:proofs}), which order the compositions in the same way.

\begin{proposition}[States for a composed update]
\label{prop:composition}
(a) Let the corridor hold any rune sequence at any positions, let the agent's state follow $m_t=f(m_{t-1},o_t)$ and its action be a function of $m_t$, and let the agent select the correct branch after every cue and every rune sequence in $\Sigma^\ast$. Then at least $Q^\star(\Sigma)$ memory values are reachable at a single corridor position. (b) If the target is a function of the cue and the rune state, then $Q^\star(\Sigma)\le K\,Q(\Sigma)$, and the deferred solution, which stores the cue and the rune state separately, uses exactly $K\,Q(\Sigma)$ states.
\end{proposition}

\autoref{tab:update-classes} (right) lists both counts, which we compute by minimizing the automata defined by the rune semantics (\autoref{app:proofs}). The counts have two uses. First, $Q^\star$ ranges from $2$ for \cInv{} to $10$ for \cRSI{}, and FFM, SHM, and GTrXL reach their lowest success rates on compositions with $Q^\star\ge4$ (\autoref{sec:results}). Second, the gap between $Q^\star$ and $KQ$ separates two kinds of solutions. An online solution folds each rune into the current target and needs $Q^\star$ states. A deferred solution keeps the cue apart from the rune state and needs $KQ$ states. For \cRes{}, $Q^\star=2Q$, because a reset makes the initial cue the target again and every solution has to keep the cue. In Rune Multi-Corridor, \cInv{} permutes the $K$ targets, so the online solution needs $K$ states and the deferred solution $2K$.

\section{ALER}
\label{sec:aler}
ALER pairs a recurrent state with a slot memory, following the two-memory view studied in RL~\citep{yang2023twomemory}, and assigns the requirements of \autoref{sec:theory} to different components. The LSTM processes every observation and can track the rune state. For retention, a slot keeps its content while later writes put little total weight on it. For rewriting, independently addressed writes overwrite one slot in a single step. For fusion, a learned gate combines the retrieved content with the recurrent state at every step. PPO~\citep{schulman2017ppo} trains all parts jointly.

\begin{figure}[t]
  \centering
  \includegraphics[width=1\textwidth]{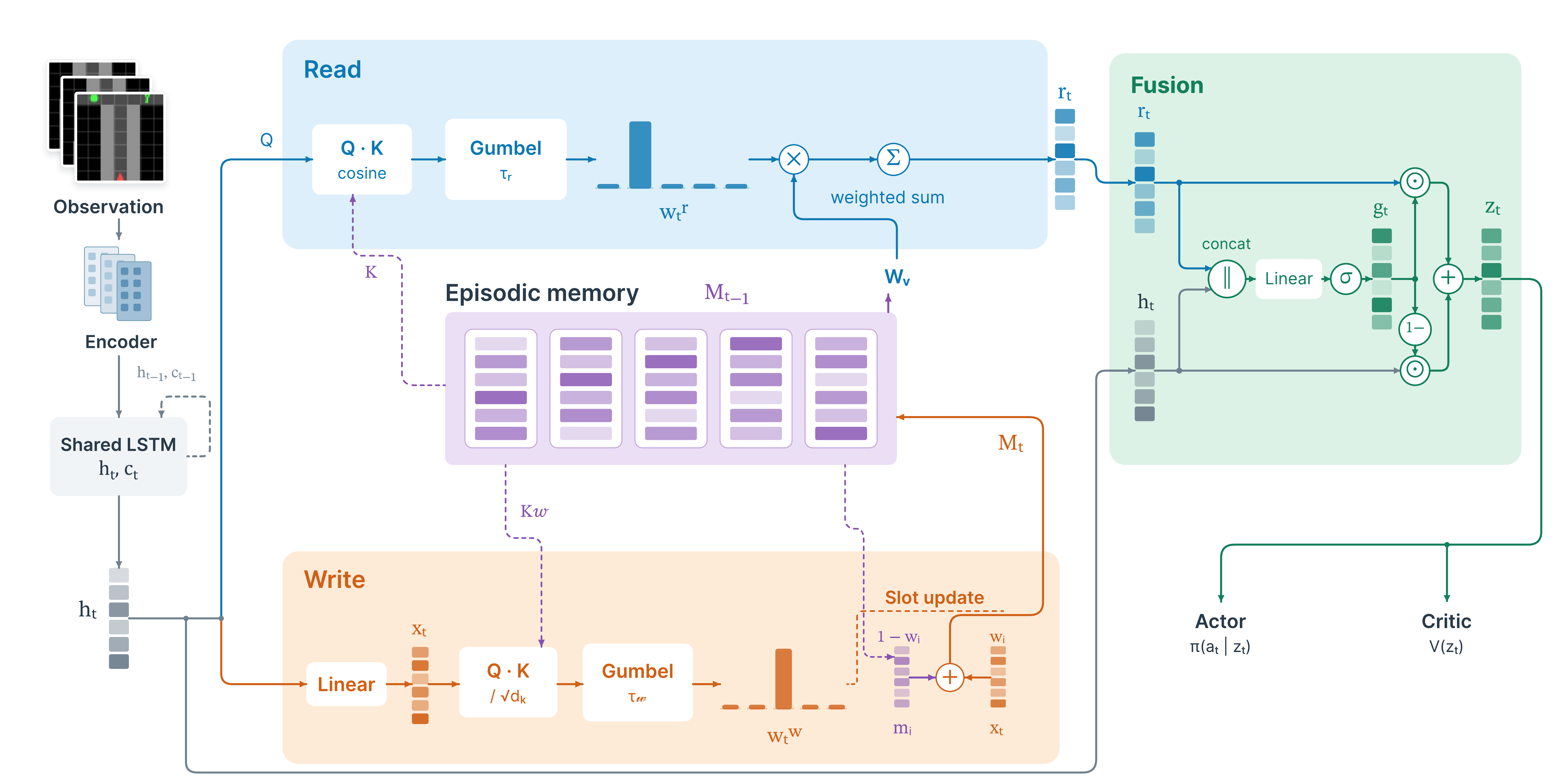}
  \vspace{-1em}
  \caption{\textbf{ALER architecture.} The shared LSTM supplies the read query, the recurrent input to fusion, and the write candidate. Reading uses $M_{t-1}$, and independent writes produce $M_t$.}
  \label{fig:aler-architecture}
  \vspace{-1em}
\end{figure}

\paragraph{One step.}
At step $t$ (\autoref{fig:aler-architecture}), the LSTM receives only the observation features, $(h_t,c_t)=\mathrm{LSTM}(\phi(o_t),h_{t-1},c_{t-1})$. The recurrent state $h_t$ then queries the memory $M_{t-1}\in\mathbb{R}^{C\times d}$ with slots $m_{t-1,i}\in\mathbb{R}^{d}$, the gate combines $h_t$ with the retrieved vector $r_t$ into the fused representation $z_t$, the actor and critic heads both receive $z_t$, and a write updates the memory to $M_t$. The retrieved vector enters only the fusion step, so the memory stores projections of past recurrent states and the LSTM computes its update from observations alone. The read uses a query, key, and value projection,
\begin{equation}
\begin{aligned}
q_t^r &= \mathrm{normalize}(W_Qh_t), \qquad
k_{t,i}^r = \mathrm{normalize}(W_Km_{t-1,i}),\\
w_t^r &= \mathrm{GumbelSoftmax}\!\left(
    [(q_t^r)^\top k_{t,i}^r]_{i=1}^{C},\tau_r\right), \qquad
r_t = \sum_{i=1}^{C}w_{t,i}^r W_Vm_{t-1,i}.
\end{aligned}
\label{eq:aler-read}
\end{equation}
A learned gate controls the contribution of each retrieved feature,
\begin{equation}
g_t=\sigma(W_g[h_t,r_t]+b_g), \qquad
z_t=g_t\odot r_t+(1-g_t)\odot h_t,
\label{eq:aler-fusion}
\end{equation}
where $\sigma$ is the sigmoid, $[\cdot,\cdot]$ denotes concatenation, and $\odot$ is elementwise multiplication. The write candidate $x_t=W_{\mathrm{write}}h_t+b_{\mathrm{write}}$ is compared with the slots through separate projections of dimension $d_k$,
\begin{equation}
\begin{aligned}
\ell_{t,i}^w &=
    \frac{(W_{Q_w}x_t)^\top(W_{K_w}m_{t-1,i})}{\sqrt{d_k}},
    \qquad w_t^w=\mathrm{GumbelSoftmax}(\ell_t^w,\tau_w),\\
m_{t,i} &= (1-w_{t,i}^w)m_{t-1,i}+w_{t,i}^w x_t .
\end{aligned}
\label{eq:aler-write}
\end{equation}

\paragraph{Addressing and initialization.}
Both addressing steps use $\mathrm{GumbelSoftmax}(\ell,\tau)=\mathrm{softmax}((\ell+\gamma)/\tau)$ with Gumbel noise $\gamma$~\citep{jang2017categoricalreparameterizationgumbelsoftmax}. The weights stay soft in training and evaluation, gradients flow through the relaxed weights, and fresh noise is drawn at every step, separately for reading and writing. As the temperature decreases, the relaxed samples approach one-hot vectors~\citep{jang2017categoricalreparameterizationgumbelsoftmax}. The temperatures are fixed, with $\tau_r=0.1$ and $\tau_w\in\{0.1,0.2\}$ in the vector environments. At evaluation, actions are greedy. The memory and the LSTM state are zero at the start of every episode. We use $C=10$ slots and $d=d_k=256$ there, and \autoref{app:hyper} lists the MiniGrid values. \autoref{app:aler-detailed-description} gives the pseudocode and the first step of an episode.

\paragraph{Properties of the write operation.}
Write weights lie on the simplex, $w_t^w\in\Delta^{C-1}$, in training and evaluation, which gives the write three properties.

\begin{proposition}[Soft overwrite]
\label{prop:write}
Let $M_0=0$. For every slot $i$ and step $t\ge1$,
(a) $m_{t,i}=\sum_{s\le t} w^w_{s,i}\prod_{u=s+1}^{t}(1-w^w_{u,i})\,x_s$,
(b) $\|m_{t,i}\|\le\max_{s\le t}\|x_s\|$, and
(c) if $w^w_{t,i}\ge1-\varepsilon$, then $\|m_{t,i}-x_t\|\le\varepsilon\|m_{t-1,i}-x_t\|$ and $\sum_{j\ne i}w^w_{t,j}\le\varepsilon$.
\end{proposition}

By (a), content written at step $s$ decays only through the write weights that later steps assign to the same slot. By (b), slot norms stay bounded at every horizon, because $x_s$ is an affine map of the bounded LSTM output. By (c), one write with weight close to one overwrites a slot, and every other slot $j$ moves toward $x_t$ by the fraction $w^w_{t,j}$, with $\sum_{j\ne i}w^w_{t,j}\le\varepsilon$. The constraint $\sum_i w^w_{t,i}=1$ fixes the total update weight across slots to one, while addressing determines how this weight is distributed. In the trained agent, one write raises the decoding of a new cue from the memory from $0.52$ to $0.99$ (\autoref{sec:mechanism}), so a single write at a cue step stores the cue.

\section{The \runemazes{} Environments}
\label{sec:environments}
\begin{figure*}[t]
  \centering
  \includegraphics[width=0.70\textwidth]{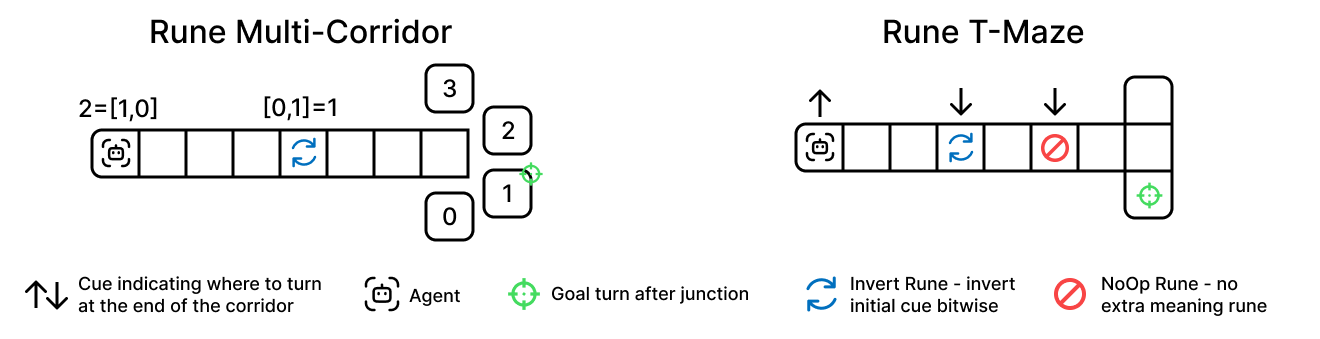}\\[4pt]
  \includegraphics[width=\textwidth]{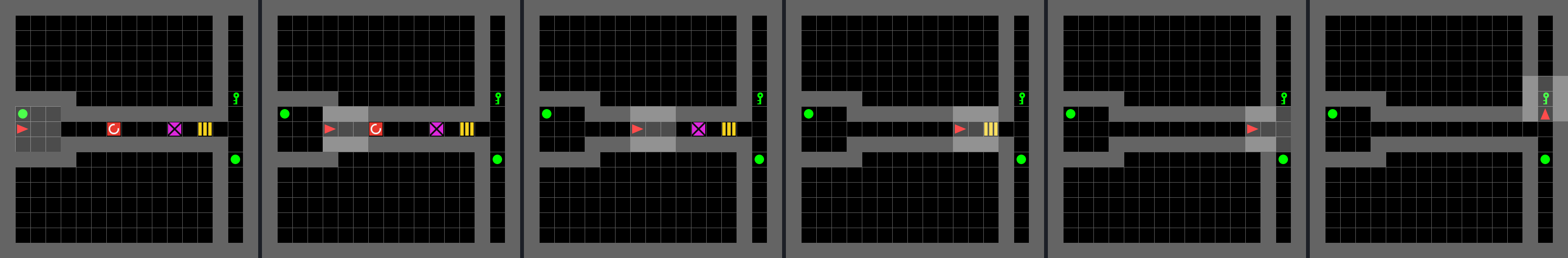}
  \caption{\textbf{\runemazes{} environments.}
  \textbf{Top left:} Rune Multi-Corridor, in which a multi-bit cue indicates the target branch and an \cInv{} rune before the junction changes it.
  \textbf{Top right:} Rune T-Maze, in which the agent must invert the initial cue after an \cInv{} rune and ignore a \texttt{NoOp} rune.
  \textbf{Bottom:} Rune MiniGrid Memory with \cRSI{}, one episode from the cue (left) to the junction (right). The agent crosses \texttt{RepeatPrev} (red), \texttt{Invert} (purple), and \texttt{SkipNext} (yellow), and each rune fires once on contact (legend in \autoref{tab:rune-legend}).}
  \vspace{-1em}
  \label{fig:rune-mazes}
  \vspace{-0.5em}
\end{figure*}

\runemazes{} instantiate the update classes of \autoref{tab:update-classes} in corridors that end in a junction. Together with Endless T-Maze~\citep{shchendrigin2026memory}, in which each of $n$ consecutive corridors shows a new cue and requires rewriting, they cover all three classes. \autoref{fig:rune-mazes} shows the vector environments.

\paragraph{Rune T-Maze.}
The agent walks a corridor of length $10$ and turns left or right at the junction. The first observation shows a binary cue, which is masked afterwards. Each rune of the composition appears at a random corridor position with probability $0.8$ and acts on the target as in \autoref{tab:update-classes}. The observation contains a junction indicator, the cue, a direction flag, and the identifier of the rune just crossed. The corridor coordinate is not observed, so the agent has to carry the cue and the rule state to the junction itself. A correct turn gives $+1$, a wrong turn $-1$, and every other step $-0.01$. We use the five compositions of \autoref{tab:update-classes}, whose $Q^\star$ ranges from $2$ to $10$ (\autoref{app:rune-t-maze}).

\paragraph{Rune Multi-Corridor.}
This environment raises the size of the target. The junction has $K\in\{4,8,16\}$ branches, and the first observation encodes the target branch as a cue of $\lceil\log_2K\rceil$ bits, which is masked afterwards. In the \cInv{} variant, one rune appears with probability $0.8$ and inverts all bits of the target, $y\leftarrow K-1-y$. Observations and rewards follow Rune T-Maze (\autoref{app:rune-multicorridor}).

\paragraph{Rune MiniGrid Memory.}
This environment tests the same rune semantics under pixel observations. It embeds the runes of Rune T-Maze into MiniGrid Memory~\citep{chevalierboisvert2023minigridminiworldmodular}, where the agent sees a cue object in a start room and has to reach the matching object at the end of a corridor. Runes are colored floor tiles on corridor cells that every path crosses, placed after the cue room and at least two cells apart. A rune may be visible before contact, fires on entry, and then disappears (\autoref{fig:rune-mazes} and \autoref{fig:rune-minigrid-rollouts}). We use the five Rune T-Maze compositions with fixed and random corridor lengths and keep MiniGrid's actions, observations, and rewards (\autoref{app:rune-minigrid}).

\section{Experimental Setup}
\label{sec:setup}
\paragraph{Baselines.}
We compare ALER with seven architectures that cover the main approaches to memory in partially observable RL: PPO-LSTM~\citep{schulman2017ppo} (recurrent state), GTrXL~\citep{parisotto2020stabilizing} (attention over cached activations), Resettable S5~\citep{lu2023structuredstatespace} (structured state-space recurrence), FFM~\citep{morad2023reinforcement} and SHM~\citep{le2025stable} (structured matrix memory), DNC~\citep{graves2016dnc} in the POPGym implementation~\citep{morad2023popgymbenchmarkingpartiallyobservable} (external memory with erase and add writes), and the memory-free PPO-MLP. On Rune MiniGrid Memory, our pixel pipeline trains ALER and PPO-LSTM, the strongest baseline on Rune Multi-Corridor, with the same CNN encoder, LSTM size, and PPO settings (\autoref{app:hyper}).

\paragraph{Training and model selection.}
For ALER, S5, and DNC on Endless T-Maze and for all architectures on Rune T-Maze and Rune Multi-Corridor, we train five agents per configuration with training seeds $42$--$46$. Rune MiniGrid Memory and the analysis in \autoref{sec:mechanism} use three training seeds. Every $10^5$ environment steps, we evaluate the current checkpoint on $100$ validation episodes generated with seed $123$ and select the checkpoint with the highest success rate, breaking ties by mean return. All result tables report the success rate of this \emph{selected checkpoint} as mean $\pm$ standard error (SEM) over training seeds. The selection rule is the same for all eight architectures. The analysis in \autoref{sec:mechanism} uses episodes independent of checkpoint selection: evaluation seeds $1$--$3$ with $100$ episodes each for gate experiments, and $128$ further episodes per checkpoint with an episode-disjoint $75/25$ split for probes.

\paragraph{Hyperparameters and capacity.}
On Endless T-Maze, PPO-LSTM, GTrXL, SHM, FFM, and PPO-MLP use the architecture-specific configurations published by \citet{shchendrigin2026memory}. For ALER and S5 on Endless T-Maze and for all architectures on Rune T-Maze and Rune Multi-Corridor, we first sweep architecture parameters and then PPO parameters with two million environment steps per candidate. We keep one hyperparameter setting per model and environment family, selected on representative configurations of that family, and use it for every rune composition and branch count (\autoref{app:hyper}). On the vector tasks, ALER has about $1.1$M parameters, about $0.5$M of them in the memory module, and PPO-LSTM about $2.3$M.

\paragraph{Research questions.}
Each rune composition and branch count is a separate training task, so the experiments measure whether an agent learns a composed update rule from reward.
\textbf{RQ1} (rewriting): does ALER replace obsolete cues in Endless T-Maze, at the training horizon and beyond it?
\textbf{RQ2} (fusion): does ALER learn composed updates of a binary cue in Rune T-Maze and of a cue with $K\in\{4,8,16\}$ values in Rune Multi-Corridor?
\textbf{RQ3} (fusion from pixels): does ALER learn the same rune compositions from pixel observations in Rune MiniGrid Memory?
\textbf{RQ4} (mechanism): where does ALER keep the cue and the rule state, and how does its policy use the memory read?
\autoref{sec:results} answers RQ1--RQ3, and \autoref{sec:mechanism} answers RQ4.

\begin{wrapfigure}{r}{0.40\textwidth}
  \vspace{-4.5em}
  \centering
  \includegraphics[width=\linewidth]{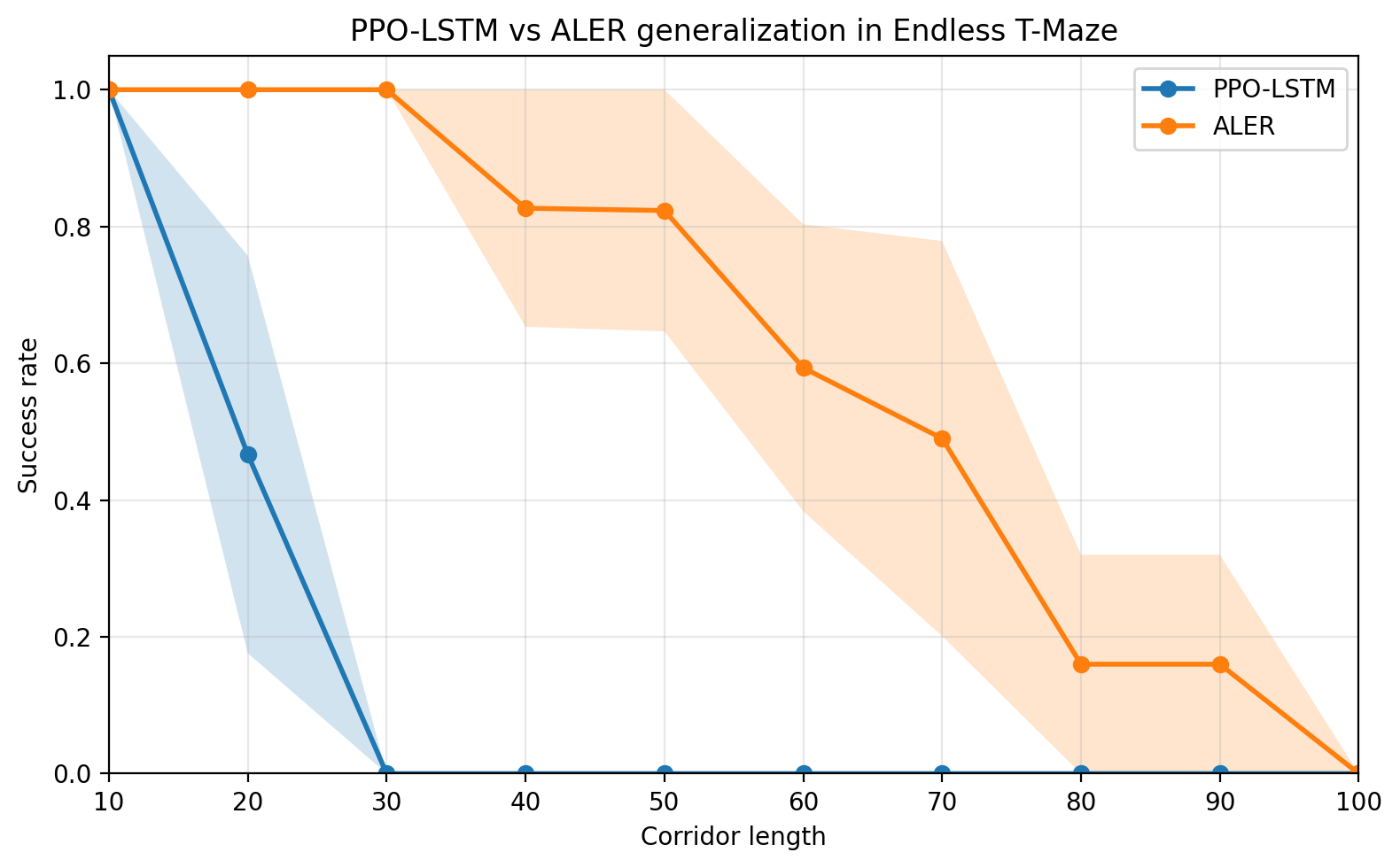}
  \caption{\textbf{Endless T-Maze horizon extrapolation.} Mean success rate over five training seeds for agents trained on $10$ corridors of length $10$ and evaluated on $10$ corridors of lengths $10$ to $100$. Shaded regions show the SEM.}
  \vspace{-1em}
  \label{fig:endless-generalization}
  \vspace{-1em}
\end{wrapfigure}

\section{Results}
\label{sec:results}
\paragraph{RQ1: rewriting in Endless T-Maze.}
ALER replaces obsolete cues reliably at the training horizon. It reaches a success rate of $1.00$ in fifteen of the sixteen Endless T-Maze configurations and $0.82\pm0.17$ in the remaining one (\autoref{tab:baselines} for constant corridor lengths and \autoref{tab:endless-uniform} in \autoref{app:endless-uniform} for random ones). ALER and S5 are the only architectures with a success rate of at least $0.80$ in every configuration. PPO-LSTM reaches $1.00$ in thirteen configurations and $0.69\pm0.31$ to $0.98\pm0.02$ in the other three. The success of GTrXL, SHM, FFM, and PPO-MLP generally decreases as the number of corridors, and with it the number of rewrites, grows.

\begin{table}[t]
\centering
\small
\caption{\textbf{Endless T-Maze.} SR of the selected checkpoint on the $100$ validation episodes, mean $\pm$ SEM over five training seeds. A configuration has $n$ corridors of constant length $l$ (\autoref{tab:endless-uniform} reports corridor lengths sampled per corridor). $n{=}1$ is the classic T-Maze and requires retention only, and every further corridor requires one rewrite. Bold marks every value equal to the best mean in its row.}
\vspace{-1em}
\label{tab:baselines}
\begin{adjustbox}{width=\textwidth,center}
\begin{tabular}{ccccccccccc}
\toprule
\textbf{$l$} & \textbf{$n$} & \textbf{PPO-MLP} & \textbf{PPO-LSTM} & \textbf{GTrXL} & \textbf{S5} & \textbf{FFM} & \textbf{SHM} & \textbf{DNC} & \textbf{ALER} \\
\midrule
5 & 1 & $0.50\pm0.00$ & $\mathbf{1.00\pm0.00}$ & $0.50\pm0.19$ & $0.80\pm0.12$ & $\mathbf{1.00\pm0.00}$ & $\mathbf{1.00\pm0.00}$ & $\mathbf{1.00\pm0.00}$ & $\mathbf{1.00\pm0.00}$ \\
5 & 3 & $0.24\pm0.05$ & $\mathbf{1.00\pm0.00}$ & $0.17\pm0.06$ & $\mathbf{1.00\pm0.00}$ & $\mathbf{1.00\pm0.00}$ & $0.67\pm0.24$ & $0.00\pm0.00$ & $\mathbf{1.00\pm0.00}$ \\
5 & 5 & $0.05\pm0.02$ & $\mathbf{1.00\pm0.00}$ & $0.04\pm0.02$ & $\mathbf{1.00\pm0.00}$ & $\mathbf{1.00\pm0.00}$ & $0.84\pm0.13$ & $0.00\pm0.00$ & $\mathbf{1.00\pm0.00}$ \\
5 & 10 & $0.00\pm0.00$ & $\mathbf{1.00\pm0.00}$ & $0.61\pm0.17$ & $\mathbf{1.00\pm0.00}$ & $\mathbf{1.00\pm0.00}$ & $0.23\pm0.22$ & $0.00\pm0.00$ & $\mathbf{1.00\pm0.00}$ \\
10 & 1 & $0.50\pm0.00$ & $\mathbf{1.00\pm0.00}$ & $0.43\pm0.01$ & $0.80\pm0.12$ & $\mathbf{1.00\pm0.00}$ & $0.84\pm0.16$ & $\mathbf{1.00\pm0.00}$ & $\mathbf{1.00\pm0.00}$ \\
10 & 3 & $0.16\pm0.02$ & $\mathbf{1.00\pm0.00}$ & $0.17\pm0.01$ & $\mathbf{1.00\pm0.00}$ & $\mathbf{1.00\pm0.00}$ & $0.93\pm0.07$ & $0.00\pm0.00$ & $0.82\pm0.17$ \\
10 & 5 & $0.04\pm0.00$ & $0.69\pm0.31$ & $0.19\pm0.12$ & $\mathbf{1.00\pm0.00}$ & $\mathbf{1.00\pm0.00}$ & $0.52\pm0.24$ & $0.00\pm0.00$ & $\mathbf{1.00\pm0.00}$ \\
10 & 10 & $0.00\pm0.00$ & $0.83\pm0.17$ & $0.02\pm0.02$ & $\mathbf{1.00\pm0.00}$ & $0.73\pm0.27$ & $0.02\pm0.01$ & $0.00\pm0.00$ & $\mathbf{1.00\pm0.00}$ \\
\bottomrule
\end{tabular}
\end{adjustbox}
\vspace{-1em}
\end{table}

ALER also keeps its success on corridors longer than those seen in training. Both agents in \autoref{fig:endless-generalization} are trained on $10$ corridors of length $10$ and evaluated on $10$ corridors of lengths $10$ to $100$, which changes the horizon and keeps the task semantics. ALER stays at a success rate of $1.00$ up to length $30$, three times the training length, and PPO-LSTM falls below $0.5$ at length $20$ and reaches zero at length $30$. \autoref{prop:write}(a) is consistent with this gap, since an earlier write in a slot decays by $\prod_u(1-w^w_{u,i})\ge1-\sum_u w^w_{u,i}$, a bound set by the total later write weight on the slot at any corridor length.

\paragraph{RQ2: fusion in Rune T-Maze.}
ALER learns all five composed updates. It reaches $1.00$ on \cInv{}, \cNoInv{}, and \cSkInv{} and $0.99$ on \cRes{} and \cRSI{} (\autoref{tab:rune-tmaze-results}). On $300$ episodes from evaluation seeds $1$--$3$, unused in checkpoint selection, the ALER agents of \autoref{sec:mechanism} reach $1.00\pm0.00$ on \cInv{} and \cRSI{} (\autoref{tab:aler-eval-ablations}). ALER and S5 are the only architectures with a success rate of at least $0.99$ on all five compositions. GTrXL reaches at least $0.99$ on four compositions and $0.92\pm0.08$ on \cRSI{}, the composition with the largest $Q^\star$. For FFM and SHM, success follows the counts of \autoref{prop:composition}. Each has a higher success rate on both compositions with $Q^\star_{\mathrm{ep}}=2$ than on any composition with $Q^\star_{\mathrm{ep}}\ge4$. PPO-MLP stays between $0.41$ and $0.51$, close to the bound of $0.5$ that \autoref{prop:bound} gives for a memoryless policy.

\begin{table}[t]
\centering
\small
\caption{\textbf{Rune T-Maze.} Success rate of the selected checkpoint on the $100$ validation episodes, mean $\pm$ SEM over five training seeds. Compositions are ordered by $Q^\star$ and $Q^\star_{\mathrm{ep}}$, which give the same order (\autoref{tab:update-classes}). Bold marks every value equal to the best mean in its row.}
\vspace{-1em}
\label{tab:rune-tmaze-results}
\begin{adjustbox}{width=\textwidth,center}
\begin{tabular}{lccccccccc}
\toprule
\textbf{Composition} & $Q^\star$ & \textbf{PPO-MLP} & \textbf{PPO-LSTM} & \textbf{GTrXL} & \textbf{S5} & \textbf{FFM} & \textbf{SHM} & \textbf{DNC} & \textbf{ALER} \\
\midrule
\cInv & 2 & $0.42\pm0.11$ & $0.84\pm0.04$ & $\mathbf{1.00\pm0.00}$ & $\mathbf{1.00\pm0.00}$ & $\mathbf{1.00\pm0.00}$ & $0.72\pm0.09$ & $\mathbf{1.00\pm0.00}$ & $\mathbf{1.00\pm0.00}$ \\
\cNoInv & 2 & $0.51\pm0.00$ & $0.78\pm0.00$ & $\mathbf{1.00\pm0.00}$ & $\mathbf{1.00\pm0.00}$ & $\mathbf{1.00\pm0.00}$ & $0.74\pm0.06$ & $0.48\pm0.07$ & $\mathbf{1.00\pm0.00}$ \\
\cSkInv & 4 & $0.51\pm0.01$ & $\mathbf{1.00\pm0.00}$ & $\mathbf{1.00\pm0.00}$ & $\mathbf{1.00\pm0.00}$ & $0.65\pm0.08$ & $0.53\pm0.00$ & $0.56\pm0.02$ & $\mathbf{1.00\pm0.00}$ \\
\cRes & 4 & $0.42\pm0.11$ & $0.81\pm0.05$ & $\mathbf{0.99\pm0.00}$ & $\mathbf{0.99\pm0.00}$ & $0.59\pm0.03$ & $0.51\pm0.02$ & $0.48\pm0.11$ & $\mathbf{0.99\pm0.00}$ \\
\cRSI & 10 & $0.41\pm0.10$ & $0.83\pm0.02$ & $0.92\pm0.08$ & $\mathbf{0.99\pm0.01}$ & $0.62\pm0.01$ & $0.55\pm0.01$ & $0.50\pm0.00$ & $\mathbf{0.99\pm0.01}$ \\
\bottomrule
\end{tabular}
\end{adjustbox}
\end{table}

\begin{table*}[t]
\centering
\small
\caption{\textbf{Rune Multi-Corridor.} SR of the selected checkpoint on the $100$ validation episodes, mean $\pm$ SEM over five training seeds. $K$ branches correspond to cues of $2$, $3$, and $4$ bits. The no-rune variant requires retention of the cue, and the \cInv{} variant requires fusion. \textbf{Random} is the success rate of a uniform guess, $1/K$. Bold marks every value equal to the best mean in its row.}
\vspace{-1em}
\label{tab:rune_multi_results}
\begin{adjustbox}{width=\textwidth,center}
\begin{tabular}{llccccccccc}
\toprule
\textbf{Setting} & $K$ & \textbf{PPO-MLP} & \textbf{PPO-LSTM} & \textbf{GTrXL} & \textbf{S5} & \textbf{FFM} & \textbf{SHM} & \textbf{DNC} & \textbf{ALER} & \textbf{Random} \\
\midrule
\multirow{3}{*}{\textbf{No runes}}
& 4 & $0.00\pm0.00$ & $\mathbf{1.00\pm0.00}$ & $0.00\pm0.00$ & $0.55\pm0.14$ & $0.00\pm0.00$ & $0.00\pm0.00$ & $0.00\pm0.00$ & $\mathbf{1.00\pm0.00}$ & $0.25$ \\
& 8 & $\mathbf{0.03\pm0.03}$ & $0.00\pm0.00$ & $0.00\pm0.00$ & $0.00\pm0.00$ & $0.00\pm0.00$ & $\mathbf{0.03\pm0.03}$ & $0.00\pm0.00$ & $0.00\pm0.00$ & $0.125$ \\
& 16 & $0.02\pm0.02$ & $0.00\pm0.00$ & $0.00\pm0.00$ & $0.00\pm0.00$ & $0.01\pm0.01$ & $0.00\pm0.00$ & $0.08\pm0.01$ & $\mathbf{0.10\pm0.03}$ & $0.0625$ \\
\midrule
\multirow{3}{*}{\textbf{\cInv}}
& 4 & $0.05\pm0.05$ & $0.77\pm0.00$ & $0.00\pm0.00$ & $0.55\pm0.16$ & $0.00\pm0.00$ & $0.00\pm0.00$ & $0.21\pm0.02$ & $\mathbf{0.83\pm0.17}$ & $0.25$ \\
& 8 & $0.02\pm0.02$ & $\mathbf{0.35\pm0.10}$ & $0.00\pm0.00$ & $0.12\pm0.06$ & $0.03\pm0.03$ & $0.06\pm0.04$ & $0.16\pm0.04$ & $0.00\pm0.00$ & $0.125$ \\
& 16 & $0.00\pm0.00$ & $\mathbf{0.10\pm0.03}$ & $0.00\pm0.00$ & $0.07\pm0.04$ & $0.02\pm0.01$ & $0.00\pm0.00$ & $0.05\pm0.02$ & $0.00\pm0.00$ & $0.0625$ \\
\bottomrule
\end{tabular}
\end{adjustbox}
\vspace{-1em}
\end{table*}

\begin{wraptable}{r}{0.5\textwidth}
  \vspace{-14pt}
  \footnotesize
  \centering
  \caption{\textbf{Rune MiniGrid Memory.} Success rate of the selected checkpoint within $30$M steps on the $100$ validation episodes, mean $\pm$ STD over three training seeds. Bold marks the higher mean. Rows are compositions with \texttt{I}\,=\,\texttt{Invert}, \texttt{N}\,=\,\texttt{NoOp}, \texttt{S}\,=\,\texttt{SkipNext}, \texttt{R}\,=\,\texttt{ResetRules}, and \texttt{P}\,=\,\texttt{RepeatPrev}.}
  \label{tab:rune-minigrid-results}
  \setlength{\tabcolsep}{2pt}
  \resizebox{\linewidth}{!}{%
  \begin{tabular}{@{}lcccc@{}}
    \toprule
    & \multicolumn{2}{c}{\textbf{Fixed length}} & \multicolumn{2}{c}{\textbf{Random length}} \\
    \cmidrule(lr){2-3}\cmidrule(lr){4-5}
    & \textbf{PPO-LSTM} & \textbf{ALER} & \textbf{PPO-LSTM} & \textbf{ALER} \\
    \midrule
    \texttt{I} & $\mathbf{0.80{\pm}0.03}$ & $0.69{\pm}0.09$ & $0.93{\pm}0.07$ & $\mathbf{0.94{\pm}0.06}$ \\
    \texttt{N+I} & $\mathbf{0.94{\pm}0.06}$ & $0.77{\pm}0.03$ & $0.67{\pm}0.07$ & $\mathbf{0.92{\pm}0.08}$ \\
    \texttt{S+I} & $0.51{\pm}0.03$ & $\mathbf{0.59{\pm}0.01}$ & $0.62{\pm}0.07$ & $\mathbf{0.63{\pm}0.05}$ \\
    \texttt{3I+R} & $0.53{\pm}0.04$ & $\mathbf{0.57{\pm}0.01}$ & $0.53{\pm}0.01$ & $\mathbf{0.57{\pm}0.01}$ \\
    \texttt{P+S+I} & $0.49{\pm}0.03$ & $\mathbf{0.60{\pm}0.00}$ & $0.56{\pm}0.00$ & $\mathbf{0.69{\pm}0.06}$ \\
    \bottomrule
  \end{tabular}}
  \vspace{-8pt}
\end{wraptable}
\paragraph{RQ2: fusion in Rune Multi-Corridor.}
At $K=4$ without runes, ALER and PPO-LSTM share the highest success rate, $1.00$, and S5 reaches $0.55\pm0.14$ (\autoref{tab:rune_multi_results}). With one \cInv{} rune, ALER has the highest mean, $0.83\pm0.17$, against $0.77\pm0.00$ for PPO-LSTM and $0.55\pm0.16$ for S5. At $K\ge8$, where the cue needs $\log_2K$ bits (\autoref{prop:bound}), no architecture exceeds $0.35$. There, a truncated episode returns $-0.20$ and a uniform guess $2/K-1$ before step penalties, $-0.75$ at $K=8$ and $-0.875$ at $K=16$ (\autoref{app:rune-multicorridor}), so an agent that has not decoded the cue earns more by never selecting a branch. Across the three vector environments, ALER is the only architecture that reaches at least $0.82$ in every Endless T-Maze configuration, at least $0.99$ on every Rune T-Maze composition, and the highest mean at $K=4$ with and without the \cInv{} rune.

\paragraph{RQ3: fusion from pixels in Rune MiniGrid Memory.}
From pixel observations on Rune MiniGrid Memory, ALER has a higher mean success rate than PPO-LSTM in eight of ten configurations and in all five random-length variants, $0.70$ against $0.66$ on average (\autoref{tab:rune-minigrid-results}). Its largest lead is on random-length \cNoInv{}, $0.92\pm0.08$ against $0.67\pm0.07$. On fixed-length corridors, ALER leads on the three compositions with $Q^\star\ge4$.

\section{What ALER Stores and How It Uses Memory}
\label{sec:mechanism}
Linear probes, scored by balanced accuracy on held-out episodes, and evaluation-time experiments on the gate test three mechanisms from Propositions~\ref{prop:composition} and~\ref{prop:write}: one write stores a new cue, stored content persists, and a deferred solution keeps the cue apart from the rune state (\autoref{app:ablation-details}).

\paragraph{The policy needs both the recurrent state and the read.}
We ablate the read path and the learned fusion at evaluation. On Endless T-Maze and on Rune T-Maze \cInv{} and \cRSI{}, we replace the learned gate by a constant $g_t$, with frozen weights and active memory writes. The constant $g_t=0$ removes the read path and passes $h_t$ alone, $g_t=0.5$ replaces learned fusion by a fixed equal mixture, and $g_t=1$ passes $r_t$ alone. Learned fusion has the highest mean success rate on all three tasks (\autoref{tab:aler-eval-ablations}), and a constant $g_t=1$ gives at most $0.26$. Setting $g_t=0$ lowers success by $0.24$ on Endless T-Maze with $20$ corridors of length $50$, beyond the training range, and by $0.17$ and $0.28$ on Rune T-Maze \cInv{} and \cRSI{}.

\paragraph{One write stores a new cue, and the cue stays decodable.}
In Endless T-Maze at length $50$, the first observation of a new cue raises its decoding from $0.52\pm0.01$ in the memory before the write, $M_t^-$, to $0.99\pm0.01$ after it, $M_t^+$, since by \autoref{prop:write}(a) the write adds $w^w_{t,i}x_t$ to every slot and $x_t$ carries the cue. Along the corridor, $M_t^+$ decodes the cue with $0.94$ to $0.99$ at all five probed positions, and $h_t$, $x_t$, and $z_t$ with $1.00$. \autoref{app:ablation-details} adds previous-cue and addressing probes.

\paragraph{The initial cue is decodable from the slot memory, and the rune flags from the recurrent state.}
On Rune T-Maze \cRSI{}, the decoding pattern of ALER matches the division of the deferred solution of \autoref{prop:composition} (\autoref{fig:aler-rune-probes}). The memory $M_t^+$ decodes the initial cue with $0.85\pm0.06$ and the \texttt{Invert} and \texttt{SkipNext} flags with $0.59\pm0.01$ and $0.47\pm0.09$, where chance is $0.5$. The recurrent state $h_t$ decodes the two flags with $0.70\pm0.02$ and $0.64\pm0.03$ and the initial cue with $0.98\pm0.01$, so the cue is available on both paths.

\paragraph{The gate reweights features at rule observations.}
If the gate carries out fusion, it should change most at rune steps. At \cInv{} rune steps in Rune T-Maze, the mean gate coefficient decreases by $0.03\pm0.01$. Compared with non-rune steps matched by position, direction, and initial cue, the mean absolute change per component, $D_t$, is higher by $0.02$ to $0.04$ at every rune type of \cInv{} and \cRSI{}, and each difference is positive in every training seed (\autoref{app:ablation-fusion}).

\section{Conclusion}
\label{sec:conclusion}
We proposed ALER, a memory-augmented RL agent, together with rune mazes, in which runes change the rule that maps an early cue to the target. To analyze these tasks, we formalized retention, rewriting, and experience fusion as three classes of target updates and derived from the rune automaton of a task how much memory a composition of rules requires. ALER assigns the three classes to separate components. Persistent slot content supports retention, independently addressed writes support rewriting, and the learned gate performs fusion, while the LSTM tracks the rule state. With this division, ALER reaches a success rate of at least $0.82$ in every Endless T-Maze configuration, keeps a success rate of $1.00$ on corridors up to three times the training length, reaches at least $0.99$ on every Rune T-Maze composition, and leads PPO-LSTM in eight of ten Rune MiniGrid Memory configurations. Probes show that the learned memory follows the deferred solution of our composition result. The slot memory stores a new cue with one write and decodes the initial cue, and the recurrent state decodes the rune flags. Rune Multi-Corridor with $K\ge8$ remains open for all eight architectures.


\subsubsection*{Reproducibility Statement}
\autoref{sec:aler} and \autoref{app:aler-detailed-description} specify ALER, including addressing noise, temperatures, and initialization. \autoref{app:rune-mazes-detailed-description} specifies the observations, actions, rewards, and rune semantics of every environment. \autoref{sec:setup} gives the training seeds, the checkpoint selection rule, and the evaluation seeds, and \autoref{app:hyper} lists all hyperparameters. 

\section*{AI Use Statement}
Generative AI tools were used solely to aid and polish the writing of the paper. Details are provided in the paper.

\begingroup
\raggedright
\bibliography{iclr2027_conference}

@inproceedings{yang2023twomemory,
  title     = {Two-Memory Reinforcement Learning},
  author    = {Yang, Zhao and Moerland, Thomas M. and Preuss, Mike and Plaat, Aske},
  booktitle = {2023 IEEE Conference on Games (CoG)},
  publisher = {IEEE},
  pages     = {1--9},
  year      = {2023},
  doi       = {10.1109/CoG57401.2023.10333174},
  eprint    = {2304.10098},
  archivePrefix = {arXiv}
}

@inproceedings{mezghani2022memory,
  title={Memory-Augmented Reinforcement Learning for Image-Goal Navigation},
  author={Mezghani, Lina and Sukhbaatar, Sainbayar and Lavril, Thibaut and Maksymets, Oleksandr and Batra, Dhruv and Bojanowski, Piotr and Alahari, Karteek},
  booktitle={2022 IEEE/RSJ International Conference on Intelligent Robots and Systems (IROS)},
  pages={3316--3323},
  year={2022}
}

@inproceedings{smith2023simplified,
  title = {Simplified State Space Layers for Sequence Modeling},
  author = {Smith, Jimmy T. H. and Warrington, Andrew and Linderman, Scott W.},
  booktitle = {International Conference on Learning Representations},
  year = {2023}
}

@inproceedings{le2025stable,
  title={Stable Hadamard Memory: Revitalizing Memory-Augmented Agents for Reinforcement Learning},
  author={Le, Hung and Nguyen, Dung and Do, Kien and Gupta, Sunil and Venkatesh, Svetha},
  booktitle={International Conference on Learning Representations (ICLR)},
  year={2025},
  note={arXiv:2410.10132}
}

@article{morad2023reinforcement,
  title={Reinforcement learning with fast and forgetful memory},
  author={Morad, Steven and Kortvelesy, Ryan and Liwicki, Stephan and Prorok, Amanda},
  journal={Advances in Neural Information Processing Systems},
  volume={36},
  pages={72008--72029},
  year={2023}
}

@inproceedings{parisotto2020stabilizing,
  title={Stabilizing transformers for reinforcement learning},
  author={Parisotto, Emilio and Song, Francis and Rae, Jack and Pascanu, Razvan and Gulcehre, Caglar and Jayakumar, Siddhant and Jaderberg, Max and Kaufman, Raphael Lopez and Clark, Aidan and Noury, Seb and others},
  booktitle={International conference on machine learning},
  pages={7487--7498},
  year={2020},
  organization={PMLR}
}

@inproceedings{shchendrigin2026memory,
  title     = {Memory Retention Is Not Enough to Master Memory Tasks in Reinforcement Learning},
  author    = {Shchendrigin, Oleg and Cherepanov, Egor and Kovalev, Alexey K. and Panov, Aleksandr I.},
  booktitle = {Proceedings of the 25th International Conference on Autonomous Agents and Multiagent Systems (AAMAS)},
  publisher = {International Foundation for Autonomous Agents and Multiagent Systems},
  year      = {2026},
  doi       = {10.65109/SOSN3643},
  eprint    = {2601.15086},
  archivePrefix = {arXiv}
}

@inproceedings{dai2019transformerxlattentivelanguagemodels,
  title     = {{Transformer-XL}: Attentive Language Models beyond a Fixed-Length Context},
  author    = {Dai, Zihang and Yang, Zhilin and Yang, Yiming and Carbonell, Jaime and Le, Quoc V. and Salakhutdinov, Ruslan},
  booktitle = {Proceedings of the 57th Annual Meeting of the Association for Computational Linguistics},
  pages     = {2978--2988},
  publisher = {Association for Computational Linguistics},
  year      = {2019},
  doi       = {10.18653/v1/P19-1285}
}

@misc{schulman2017ppo,
  title={Proximal Policy Optimization Algorithms},
  author={John Schulman and Filip Wolski and Prafulla Dhariwal and Alec Radford and Oleg Klimov},
  year={2017},
  eprint={1707.06347},
  archivePrefix={arXiv},
  primaryClass={cs.LG}
}

@article{kaelbling1998planning,
  title={Planning and acting in partially observable stochastic domains},
  author={Kaelbling, Leslie Pack and Littman, Michael L and Cassandra, Anthony R},
  journal={Artificial intelligence},
  volume={101},
  number={1-2},
  pages={99--134},
  year={1998},
  publisher={Elsevier}
}

@article{heess2015memory,
  title={Memory-based control with recurrent neural networks},
  author={Heess, Nicolas and Hunt, Jonathan J and Lillicrap, Timothy P and Silver, David},
  journal={arXiv preprint arXiv:1512.04455},
  year={2015}
}

@inproceedings{chevalierboisvert2023minigridminiworldmodular,
  title     = {{MiniGrid} \& {MiniWorld}: Modular \& Customizable Reinforcement Learning Environments for Goal-Oriented Tasks},
  author    = {Chevalier-Boisvert, Maxime and Dai, Bolun and Towers, Mark and Perez-Vicente, Rodrigo and Willems, Lucas and Lahlou, Salem and Pal, Suman and Castro, Pablo Samuel and Terry, Jordan},
  booktitle = {Advances in Neural Information Processing Systems},
  volume    = {36},
  year      = {2023},
  doi       = {10.52202/075280-3209}
}

@article{pleines2025memorygymendlesstasks,
  author  = {Marco Pleines and Matthias Pallasch and Frank Zimmer and Mike Preuss},
  title   = {Memory Gym: Towards Endless Tasks to Benchmark Memory Capabilities of Agents},
  journal = {Journal of Machine Learning Research},
  year    = {2025},
  volume  = {26},
  number  = {6},
  pages   = {1--40},
  url     = {http://jmlr.org/papers/v26/24-0043.html}
}

@inproceedings{morad2023popgymbenchmarkingpartiallyobservable,
  title     = {{POPGym}: Benchmarking Partially Observable Reinforcement Learning},
  author    = {Morad, Steven and Kortvelesy, Ryan and Bettini, Matteo and Liwicki, Stephan and Prorok, Amanda},
  booktitle = {International Conference on Learning Representations},
  year      = {2023}
}

@inproceedings{wang2025investigatingmemoryrlpopgym,
  title     = {Investigating Memory in Model-Free {RL} with {POPGym} Arcade},
  author    = {Wang, Zekang and He, Zhe and Zhang, Borong and Toledo, Edan and Morad, Steven},
  booktitle = {Proceedings of the 43rd International Conference on Machine Learning},
  pages     = {126692--126733},
  volume    = {306},
  series    = {Proceedings of Machine Learning Research},
  publisher = {PMLR},
  year      = {2026}
}

@misc{graves2014neuralturingmachines,
      title={Neural Turing Machines}, 
      author={Alex Graves and Greg Wayne and Ivo Danihelka},
      year={2014},
      eprint={1410.5401},
      archivePrefix={arXiv},
      primaryClass={cs.NE},
      url={https://arxiv.org/abs/1410.5401}, 
}

@inproceedings{pritzel2017neuralepisodiccontrol,
  title     = {Neural Episodic Control},
  author    = {Pritzel, Alexander and Uria, Benigno and Srinivasan, Sriram and Puigdom{\`e}nech Badia, Adri{\`a} and Vinyals, Oriol and Hassabis, Demis and Wierstra, Daan and Blundell, Charles},
  booktitle = {Proceedings of the 34th International Conference on Machine Learning},
  pages     = {2827--2836},
  volume    = {70},
  series    = {Proceedings of Machine Learning Research},
  publisher = {PMLR},
  year      = {2017}
}

@misc{wayne2018unsupervisedpredictivememorygoaldirected,
      title={Unsupervised Predictive Memory in a Goal-Directed Agent}, 
      author={Greg Wayne and Chia-Chun Hung and David Amos and Mehdi Mirza and Arun Ahuja and Agnieszka Grabska-Barwinska and Jack Rae and Piotr Mirowski and Joel Z. Leibo and Adam Santoro and Mevlana Gemici and Malcolm Reynolds and Tim Harley and Josh Abramson and Shakir Mohamed and Danilo Rezende and David Saxton and Adam Cain and Chloe Hillier and David Silver and Koray Kavukcuoglu and Matt Botvinick and Demis Hassabis and Timothy Lillicrap},
      year={2018},
      eprint={1803.10760},
      archivePrefix={arXiv},
      primaryClass={cs.LG},
      url={https://arxiv.org/abs/1803.10760}, 
}

@inproceedings{fortunato2020generalizationreinforcementlearnersworking,
  title     = {Generalization of Reinforcement Learners with Working and Episodic Memory},
  author    = {Fortunato, Meire and Tan, Melissa and Faulkner, Ryan and Hansen, Steven and Puigdom{\`e}nech Badia, Adri{\`a} and Buttimore, Gavin and Deck, Charles and Leibo, Joel Z. and Blundell, Charles},
  booktitle = {Advances in Neural Information Processing Systems},
  volume    = {32},
  year      = {2019}
}

@inproceedings{NEURIPS2024_b528459c,
 author = {Chateau-Laurent, Hugo and Alexandre, Fr\'{e}d\'{e}ric},
 booktitle = {Advances in Neural Information Processing Systems},
 doi = {10.52202/079017-3170},
 editor = {A. Globerson and L. Mackey and D. Belgrave and A. Fan and U. Paquet and J. Tomczak and C. Zhang},
 pages = {99921--99946},
 publisher = {Curran Associates, Inc.},
 title = {Relating Hopfield Networks to Episodic Control},
 url = {https://proceedings.neurips.cc/paper_files/paper/2024/file/b528459c99e929718a7d7e1697253d7f-Paper-Conference.pdf},
 volume = {37},
 year = {2024}
}

@inproceedings{jang2017categoricalreparameterizationgumbelsoftmax,
  title     = {Categorical Reparameterization with {Gumbel-Softmax}},
  author    = {Jang, Eric and Gu, Shixiang and Poole, Ben},
  booktitle = {International Conference on Learning Representations},
  year      = {2017}
}

@article{hochreiter1997lstm,
  title={Long Short-Term Memory},
  author={Hochreiter, Sepp and Schmidhuber, J{\"u}rgen},
  journal={Neural Computation},
  volume={9},
  number={8},
  pages={1735--1780},
  year={1997},
  publisher={MIT Press}
}

@inproceedings{ni2022recurrentstrong,
  title     = {Recurrent Model-Free {RL} Can Be a Strong Baseline for Many {POMDPs}},
  author    = {Ni, Tianwei and Eysenbach, Benjamin and Salakhutdinov, Ruslan},
  booktitle = {Proceedings of the 39th International Conference on Machine Learning},
  pages     = {16691--16723},
  volume    = {162},
  series    = {Proceedings of Machine Learning Research},
  publisher = {PMLR},
  year      = {2022}
}

@article{cayci2024recurrentnatural,
  title   = {Recurrent Natural Policy Gradient for {POMDPs}},
  author  = {Cayci, Semih and Eryilmaz, Atilla},
  journal = {Transactions on Machine Learning Research},
  year    = {2025}
}

@inproceedings{lampinen2021hcam,
  title     = {Towards Mental Time Travel: A Hierarchical Memory for Reinforcement Learning Agents},
  author    = {Lampinen, Andrew K. and Chan, Stephanie C. Y. and Banino, Andrea and Hill, Felix},
  booktitle = {Advances in Neural Information Processing Systems},
  volume    = {34},
  year      = {2021}
}

@misc{esslinger2022dtqn,
  title={Deep Transformer Q-Networks for Partially Observable Reinforcement Learning},
  author={Kevin Esslinger and Robert Platt and Christopher Amato},
  year={2022},
  eprint={2206.01078},
  archivePrefix={arXiv},
  primaryClass={cs.LG},
  url={https://arxiv.org/abs/2206.01078}
}

@inproceedings{lu2023structuredstatespace,
  title     = {Structured State Space Models for In-Context Reinforcement Learning},
  author    = {Lu, Chris and Schroecker, Yannick and Gu, Albert and Parisotto, Emilio and Foerster, Jakob and Singh, Satinder and Behbahani, Feryal},
  booktitle = {Advances in Neural Information Processing Systems},
  volume    = {36},
  pages     = {47016--47031},
  year      = {2023},
  doi       = {10.52202/075280-2036},
  eprint    = {2303.03982},
  archivePrefix = {arXiv}
}

@inproceedings{ni2023whendotransformers,
  title     = {When Do Transformers Shine in {RL}? Decoupling Memory from Credit Assignment},
  author    = {Ni, Tianwei and Ma, Michel and Eysenbach, Benjamin and Bacon, Pierre-Luc},
  booktitle = {Advances in Neural Information Processing Systems},
  volume    = {36},
  year      = {2023},
  doi       = {10.52202/075280-2194}
}

@article{graves2016dnc,
  title={Hybrid computing using a neural network with dynamic external memory},
  author={Graves, Alex and Wayne, Greg and Reynolds, Malcolm and Harley, Tim and Danihelka, Ivo and Grabska-Barwi{\'n}ska, Agnieszka and G{\'o}mez Colmenarejo, Sergio and Grefenstette, Edward and Ramalho, Tiago and Agapiou, John and Puigdom{\`e}nech Badia, Adri{\`a} and Hermann, Karl Moritz and Zwols, Yori and Ostrovski, Georg and Cain, Adam and King, Helen and Summerfield, Christopher and Blunsom, Phil and Kavukcuoglu, Koray and Hassabis, Demis},
  journal={Nature},
  volume={538},
  number={7626},
  pages={471--476},
  year={2016},
  publisher={Nature Publishing Group}
}

@article{chen2021decisiontransformer,
  title={Decision transformer: Reinforcement learning via sequence modeling},
  author={Chen, Lili and Lu, Kevin and Rajeswaran, Aravind and Lee, Kimin and Grover, Aditya and Laskin, Misha and Abbeel, Pieter and Srinivas, Aravind and Mordatch, Igor},
  journal={Advances in neural information processing systems},
  volume={34},
  pages={15084--15097},
  year={2021}
}

@inproceedings{cherepanov2023rate,
 author = {Cherepanov, Egor and Staroverov, Aleksei and Kovalev, Alexey and Panov, Aleksandr},
 booktitle = {International Conference on Learning Representations},
 editor = {C. Vondrick and B. Hariharan and C. Raffel and L. Pinto and D. Yang and A. Faust},
 pages = {129379--129407},
 title = {Recurrent Action Transformer with Memory},
 url = {https://proceedings.iclr.cc/paper_files/paper/2026/file/d21b249204cace583de6beabcb6e4b3d-Paper-Conference.pdf},
 volume = {2026},
 year = {2026}
}

@misc{tassa2018deepmindcontrolsuite,
  title={DeepMind Control Suite},
  author={Yuval Tassa and Yotam Doron and Alistair Muldal and Tom Erez and Yazhe Li and Diego de Las Casas and David Budden and Abbas Abdolmaleki and Josh Merel and Andrew Lefrancq and Timothy Lillicrap and Martin Riedmiller},
  year={2018},
  eprint={1801.00690},
  archivePrefix={arXiv},
  primaryClass={cs.AI},
  url={https://arxiv.org/abs/1801.00690}
}

@inproceedings{todorov2012mujoco,
  title={MuJoCo: A physics engine for model-based control},
  author={Todorov, Emanuel and Erez, Tom and Tassa, Yuval},
  booktitle={2012 IEEE/RSJ International Conference on Intelligent Robots and Systems},
  pages={5026--5033},
  year={2012},
  organization={IEEE}
}

@misc{beattie2016deepmindlab,
  title={DeepMind Lab},
  author={Charles Beattie and Joel Z. Leibo and Denis Teplyashin and Tom Ward and Marcus Wainwright and Heinrich K{\"u}ttler and Andrew Lefrancq and Simon Green and V{\'\i}ctor Vald{\'e}s and Amir Sadik and Julian Schrittwieser and Keith Anderson and Sarah York and Max Cant and Adam Cain and Adrian Bolton and Stephen Gaffney and Helen King and Demis Hassabis and Shane Legg and Stig Petersen},
  year={2016},
  eprint={1612.03801},
  archivePrefix={arXiv},
  primaryClass={cs.AI},
  url={https://arxiv.org/abs/1612.03801}
}

@article{bellemare2013ale,
  title={The Arcade Learning Environment: An Evaluation Platform for General Agents},
  author={Bellemare, Marc G. and Naddaf, Yavar and Veness, Joel and Bowling, Michael},
  journal={Journal of Artificial Intelligence Research},
  volume={47},
  pages={253--279},
  year={2013},
  doi={10.1613/jair.3912}
}

@inproceedings{cobbe2020procgen,
  title     = {Leveraging Procedural Generation to Benchmark Reinforcement Learning},
  author    = {Cobbe, Karl and Hesse, Chris and Hilton, Jacob and Schulman, John},
  booktitle = {Proceedings of the 37th International Conference on Machine Learning},
  pages     = {2048--2056},
  volume    = {119},
  series    = {Proceedings of Machine Learning Research},
  publisher = {PMLR},
  year      = {2020}
}

@inproceedings{juliani2019obstacletower,
  title     = {Obstacle Tower: A Generalization Challenge in Vision, Control, and Planning},
  author    = {Juliani, Arthur and Khalifa, Ahmed and Berges, Vincent-Pierre and Harper, Jonathan and Teng, Ervin and Henry, Hunter and Crespi, Adam and Togelius, Julian and Lange, Danny},
  booktitle = {Proceedings of the Twenty-Eighth International Joint Conference on Artificial Intelligence},
  pages     = {2684--2691},
  year      = {2019},
  doi       = {10.24963/ijcai.2019/373}
}

@inproceedings{pasukonis2022memorymaze,
  title     = {Evaluating Long-Term Memory in 3D Mazes},
  author    = {Pasukonis, Jurgis and Lillicrap, Timothy and Hafner, Danijar},
  booktitle = {International Conference on Learning Representations},
  year      = {2023}
}

@inproceedings{cherepanov2025mikasa,
  title     = {Memory, Benchmark \& Robots: A Benchmark for Solving Complex Tasks with Reinforcement Learning},
  author    = {Cherepanov, Egor and Kachaev, Nikita and Kovalev, Alexey K. and Panov, Aleksandr I.},
  booktitle = {International Conference on Learning Representations},
  year      = {2026}
}

@misc{gulcehre2017wormhole,
  title={Memory Augmented Neural Networks with Wormhole Connections},
  author={Gulcehre, Caglar and Chandar, Sarath and Bengio, Yoshua},
  year={2017},
  eprint={1701.08718},
  archivePrefix={arXiv},
  primaryClass={cs.LG},
  url={https://arxiv.org/abs/1701.08718}
}

@inproceedings{cherepanov2025elmur,
 author = {Cherepanov, Egor and Kovalev, Alexey and Panov, Aleksandr},
 booktitle = {International Conference on Learning Representations},
 editor = {C. Vondrick and B. Hariharan and C. Raffel and L. Pinto and D. Yang and A. Faust},
 pages = {141814--141844},
 title = {ELMUR: External Layer Memory with Update/Rewrite for Long-Horizon RL Problems},
 url = {https://proceedings.iclr.cc/paper_files/paper/2026/file/e58dfd0157cb9de91e7a513d3148aad5-Paper-Conference.pdf},
 volume = {2026},
 year = {2026}
}

@misc{zaremba2015rlntm,
  title={Reinforcement Learning Neural Turing Machines - Revised},
  author={Zaremba, Wojciech and Sutskever, Ilya},
  year={2015},
  eprint={1505.00521},
  archivePrefix={arXiv},
  primaryClass={cs.LG},
  url={https://arxiv.org/abs/1505.00521}
}

@misc{farr2026streaming,
      title={Streaming Reinforcement Learning under Partial Observability with Real-Time Recurrent Learning},
      author={Noah Farr and Aryaman Reddi and Carlo D'Eramo and Jan Peters},
      year={2026},
      eprint={2605.24709},
      archivePrefix={arXiv},
      primaryClass={cs.LG},
      url={https://arxiv.org/abs/2605.24709},
}

@inproceedings{gupta2025memo,
  title     = {{Memo}: Training Memory-Efficient Embodied Agents with Reinforcement Learning},
  author    = {Gupta, Gunshi and Yadav, Karmesh and Kira, Zsolt and Gal, Yarin and Aljundi, Rahaf},
  booktitle = {Advances in Neural Information Processing Systems},
  volume    = {38},
  year      = {2025},
  doi       = {10.52202/085713-3486}
}

@article{wu2026remember,
  title={Remember When It Matters: Proactive Memory Agent for Long-Horizon Agents},
  author={Wu, Yifan and Zhang, Lizhu and Zhou, Yuhang and Wang, Mingyi and Peng, Bo and Li, Serena and Fan, Xiangjun and Zhao, Zhuokai},
  journal={arXiv preprint arXiv:2607.08716},
  year={2026}
}

@article{liu2026worldmemarena,
  title={WorldMemArena: Evaluating Multimodal Agent Memory Through Action-World Interaction},
  author={Liu, Chengzhi and Yang, Yuzhe and Pu, Sophia Xiao and Liu, Yepeng and Long, Lin and Guo, Yichen and Chen, Nuo and Weng, Zhaotian and Kochkina, Elena and Kaur, Simerjot and Smiley, Charese and Liu, Xiaomo and Zou, James and Liu, Sheng and Bu, Yuheng and Peng, Songyou and Wang, Xin Eric},
  journal={arXiv preprint arXiv:2605.29341},
  year={2026}
}

@article{zhang2025learn,
  title={Learn to Memorize: Optimizing LLM-based Agents with Adaptive Memory Framework},
  author={Zhang, Zeyu and Dai, Quanyu and Li, Rui and Bo, Xiaohe and Chen, Xu and Dong, Zhenhua},
  journal={arXiv preprint arXiv:2508.16629},
  year={2025}
}
\bibliographystyle{iclr2027_conference}
\endgroup

\appendix
\section*{Appendix contents}
\begin{itemize}[topsep=4pt,itemsep=2pt,parsep=0pt,leftmargin=1.5em]
  \item \nameref{app:proofs} \dotfill \pageref{app:proofs}
  \item \nameref{app:aler-detailed-description} \dotfill \pageref{app:aler-detailed-description}
  \item \nameref{app:rune-mazes-detailed-description} \dotfill \pageref{app:rune-mazes-detailed-description}
  \item \nameref{app:additional-results} \dotfill \pageref{app:additional-results}
  \item \nameref{app:ablation-details} \dotfill \pageref{app:ablation-details}
  \item \nameref{app:hyper} \dotfill \pageref{app:hyper}
  \item \nameref{app:statements} \dotfill \pageref{app:statements}
\end{itemize}
\FloatBarrier

\section{Proofs}
\label{app:proofs}

\subsection{Proof of \autoref{prop:bound}}
Let the action at $\tau$ follow a possibly randomized rule $\pi(a\mid m_\tau,o_\tau)$. Then
\[
P(a_\tau=y_\tau)=\sum_{m,o,y}P(m,o,y)\,\pi(y\mid m,o)\le\sum_{m,o}\max_{y}P(m,o,y)\le\sum_{m}\sum_{o}\max_{y}P(o,y).
\]
By (D), $P(o,y)=P(o)/K$ for every $y$, so the inner sum equals $1/K$ and the total is at most $M/K$. A probability is also at most one. For continuous observations, the sum over $o$ becomes an integral with the same bound. The step $\tau$ may depend on the policy. In our environments, the agent crosses every rune before the junction, $o_\tau$ is a function of the rune placement alone, and $y_\tau$ is the cue transformed by the runes of the episode. For each placement, $y_\tau$ is a bijective function of $c$. The cue is uniform and independent of the placement, so $y_\tau$ is uniform and independent of $o_\tau$ for every policy. \qed

\subsection{Proof of \autoref{prop:composition}}
Let $T(c,\sigma)$ be the target after cue $c$ and rune sequence $\sigma\in\Sigma^\ast$. Call two prefixes $(c,\sigma)$ and $(c',\sigma')$ equivalent if $T(c,\sigma u)=T(c',\sigma' u)$ for every continuation $u\in\Sigma^\ast$. The states of the minimal automaton for $T$ are exactly these equivalence classes, so there are $Q^\star(\Sigma)$ of them.

(a) We use the corridor of the statement and index the episode by corridor position. The update $f$ and the action map are the same at every step. Take two prefixes $(c,\sigma)$ and $(c',\sigma')$, place the runes of both before a common position $p$, and place a continuation $u$ at the same positions after $p$ in both episodes. After the first step, the cue is masked, and each observation is determined by the position and the rune crossed at that step. Suppose that the memory values are equal at the first step at which the agent is at $p$ in the two episodes. From that step on, both episodes receive the same observations and take the same actions, including turns that do not move the agent, so they reach the junction the same number of steps after $p$ with the same memory and the same decision. The hypothesis requires a decision in every episode, so truncation does not occur. If the prefixes are inequivalent, some continuation $u$ gives them different targets, and the agent errs on one of the two episodes. Hence prefixes from different classes reach different memory values at $p$. Every class contains a prefix, and the corridor lets all these prefixes end before the same $p$, so at least $Q^\star(\Sigma)$ memory values are reachable at $p$. Consider next an agent whose update or action uses noise that is independent and identically distributed across steps, such as Gumbel noise, and that is correct with probability one. The first step at $p$ is a stopping time, so the noise after it is independent of the memory at $p$ and has the same law in both episodes. For inequivalent prefixes $A$ and $B$ with separating continuation $u$, let $S_A$ be the set of memory values from which the agent, given $u$, selects the target of $A$ with probability one, and define $S_B$ likewise. The memory at $p$ lies in $S_A$ with probability one in the episode of $A$ and in $S_B$ with probability one in the episode of $B$, and $S_A\cap S_B=\emptyset$ because the two targets differ. The laws of the memory at $p$ are therefore concentrated on disjoint sets for inequivalent prefixes. Take one prefix from each class. Each prefix $A$ reaches a value that lies in $S_A$ for every separating continuation, these values differ across the chosen prefixes, and at least $Q^\star(\Sigma)$ values are therefore reachable at $p$.

(b) The deferred solution keeps the cue $c$ and the state $q$ of the minimal rune automaton. Its output is the target, $c\oplus\mathrm{inv}(q)$ in Rune T-Maze and $\mathrm{inv}(q)$ applications of $y\mapsto K-1-y$ to $c$ in Rune Multi-Corridor. It uses $KQ(\Sigma)$ states and computes $T$, and the minimal automaton has at most as many states. \qed

\subsection{Computing $Q$ and $Q^\star$}
The state of the rune automaton of Rune T-Maze is $(\mathit{invert},\mathit{skip},\mathit{last})$ with $\mathit{invert},\mathit{skip}\in\{0,1\}$ and $\mathit{last}$ the last rune that took effect. The transitions follow \autoref{app:rune-t-maze}. \texttt{Invert} flips $\mathit{invert}$, \texttt{SkipNext} sets $\mathit{skip}$, \texttt{ResetRules} clears both flags, and \texttt{NoOp} changes neither. A rune that arrives while $\mathit{skip}=1$ clears $\mathit{skip}$ and leaves $\mathit{last}$ unchanged. \texttt{RepeatPrev} re-applies $\mathit{last}$ if it is a rune other than \texttt{RepeatPrev} and then sets $\mathit{last}$ to \texttt{RepeatPrev}. For each composition, we enumerate the states reachable from $(0,0,\text{none})$ over its alphabet and merge equivalent states by partition refinement. With output $\mathit{invert}$ this gives $Q$. With the cue added to the state and output $c\oplus\mathit{invert}$ it gives $Q^\star$. For \cRes{}, the cue and the invert flag are both needed after a reset, which gives $Q^\star=4=2Q$. For the other four compositions, the minimal target automaton merges the cue into the invert flag, which gives $Q^\star=Q$. A sampled episode contains each rune candidate of its composition at most once, in uniformly random order, so the rune sequences that occur form a finite set. The candidates are one \texttt{Invert} for \cInv{}, \texttt{NoOp} and \texttt{Invert} for \cNoInv{}, \texttt{SkipNext} and \texttt{Invert} for \cSkInv{}, three \texttt{Invert} and one \texttt{ResetRules} for \cRes{}, and \texttt{RepeatPrev}, \texttt{SkipNext}, and \texttt{Invert} for \cRSI{}. Each candidate is present with probability $0.8$, and every order of the present candidates fits the corridor of length $10$, so every such sequence occurs with positive probability. On this set we search for the smallest deterministic update that selects the correct branch on every sampled episode, by coloring the tree of cue and rune prefixes so that equal colors have equal outputs and equal successors. $Q^\star_{\mathrm{ep}}$ is the number of colors after the cue. $Q_{\mathrm{ep}}$ is the number of colors in the same search over rune prefixes alone with output $\mathit{invert}$, and it includes the initial state, as $Q$ does. This search gives $Q^\star_{\mathrm{ep}}=2,2,4,4,6$ and $Q_{\mathrm{ep}}=2,2,3,2,4$ for \cInv{}, \cNoInv{}, \cSkInv{}, \cRes{}, and \cRSI{}. The values of $Q^\star_{\mathrm{ep}}$ also bound the memory of an agent from below. Let $n$ be the number of candidates. For $s\le n$, let $G_s$ be the graph whose nodes are the cue and rune prefixes with at most $s$ runes and whose edges join two prefixes if a continuation of length at most $n-s$ exists that uses only candidates left unused by each of the two prefixes and gives them different targets. Rune positions have positive density in $(0.02,0.98)$ at a spacing of at least $0.1=1/L$, so each step crosses at most one rune, and observations depend on a placement only through the step at which each rune is crossed. Placing both prefixes before a common position $p$ and the continuation after $p$ therefore gives, for each of the two episodes, a set of placements of positive probability with the same crossing steps. By the argument of \autoref{prop:composition}(a), an agent that selects the correct branch with probability one needs different memory values at $p$ for adjacent prefixes, and hence at least $\chi(G_s)$ memory values. For all five compositions, the computed $\max_s\chi(G_s)$ equals $Q^\star_{\mathrm{ep}}$, so this lower bound is tight. The minimal automata over $\Sigma^\ast$ restricted to the sampled prefixes are valid colorings, so the counts over $\Sigma^\ast$ bound the sampled counts from above, and both counts order the compositions by $Q^\star$ in the same way.

\subsection{Proof of \autoref{prop:write}}
(a) By induction on $t$. The case $t=0$ holds with $M_0=0$. If (a) holds at $t-1$, substituting it into $m_{t,i}=(1-w^w_{t,i})m_{t-1,i}+w^w_{t,i}x_t$ multiplies every earlier coefficient by $(1-w^w_{t,i})$ and adds $x_t$ with coefficient $w^w_{t,i}$.
(b) The coefficients in (a) are nonnegative and sum to $1-\prod_{u=1}^{t}(1-w^w_{u,i})\le1$, so $m_{t,i}$ is a convex combination of $0,x_1,\dots,x_t$, and the triangle inequality gives the bound.
(c) $m_{t,i}-x_t=(1-w^w_{t,i})(m_{t-1,i}-x_t)$, and $1-w^w_{t,i}\le\varepsilon$. Since $w^w_t$ lies on the simplex, $\sum_{j\ne i}w^w_{t,j}=1-w^w_{t,i}\le\varepsilon$. \qed

\section{ALER Detailed Description}
\label{app:aler-detailed-description}
This section gives the pseudocode of one ALER step in the notation of \autoref{sec:aler}. The memory and the LSTM state are zero at the start of every episode, and $\mathrm{normalize}(v)=v/\max(\|v\|,10^{-8})$, so $\mathrm{normalize}(0)=0$. At the first step, all read and write logits are therefore zero, the retrieved vector is $r_1=0$, and the first write is spread over slots by the Gumbel noise alone. Algorithms~\ref{alg:aler_read}--\ref{alg:aler_write} specify the three memory operations, and \autoref{alg:aler_full} places them in the actor-critic step.

\noindent
\begin{minipage}[t]{0.32\linewidth}
\aleralgobox{Read}{alg:aler_read}{%
    \begin{algorithmic}[1]
    \Require $h_t$, $M_{t-1}$, $\tau_r$
    \State $q_t^r \gets \mathrm{normalize}(W_Q h_t)$
    \For{$i=1$ to $C$}
        \State $k_{t,i}^r \gets \mathrm{normalize}(W_K m_{t-1,i})$
        \State $s_{t,i} \gets (q_t^r)^\top k_{t,i}^r$
    \EndFor
    \State $w_t^r \gets \mathrm{GumbelSoftmax}(s_t,\tau_r)$
    \State $r_t \gets \sum_i w_{t,i}^r W_V m_{t-1,i}$
    \State \Return $r_t$
    \end{algorithmic}%
  }
\end{minipage}\hfill
\begin{minipage}[t]{0.32\linewidth}
\aleralgobox{Fusion}{alg:aler_fusion}{%
    \begin{algorithmic}[1]
    \Require $h_t$, $r_t$
    \State $g_t \gets \sigma(W_g[h_t,r_t]+b_g)$
    \State $z_t \gets g_t \odot r_t + (1-g_t) \odot h_t$
    \State \Return $z_t$
    \end{algorithmic}%
  }
\end{minipage}\hfill
\begin{minipage}[t]{0.32\linewidth}
\aleralgobox{Write}{alg:aler_write}{%
    \begin{algorithmic}[1]
    \Require $h_t$, $M_{t-1}$, $\tau_w$
    \State $x_t \gets W_{\mathrm{write}} h_t+b_{\mathrm{write}}$
    \For{$i=1$ to $C$}
        \State $\ell_{t,i}^w \gets \frac{(W_{Q_w}x_t)^\top W_{K_w}m_{t-1,i}}{\sqrt{d_k}}$
    \EndFor
    \State $w_t^w \gets \mathrm{GumbelSoftmax}(\ell_t^w,\tau_w)$
    \For{$i=1$ to $C$}
        \State $m_{t,i} \gets (1-w_{t,i}^w)m_{t-1,i} + w_{t,i}^w x_t$
    \EndFor
    \State \Return $M_t$
    \end{algorithmic}%
  }
\end{minipage}


\paragraph{Components.}
A flat feature extractor maps the observation to $\phi(o_t)$, and a single-layer LSTM with hidden size $256$ processes it. The actor and critic have separate MLP heads with two hidden layers of $256$ units, and both receive the fused representation $z_t$. The read projections are $W_Q,W_K\in\mathbb{R}^{d_k\times d}$ and $W_V\in\mathbb{R}^{d\times d}$, and the gate parameters are $W_g\in\mathbb{R}^{d\times 2d}$ and $b_g\in\mathbb{R}^{d}$. Read scores are cosine similarities between the normalized query and keys. All $C$ slots take part in reading and writing from the first step, including slots whose content is still zero.

\paragraph{Addressing in training and evaluation.}
Both addressing steps use the relaxed Gumbel-Softmax weights of \autoref{sec:aler} with noise $\gamma_i=-\log(-\log u_i)$ and $u_i\sim\mathcal{U}(10^{-6},1-10^{-6})$ in training and in evaluation, with fresh noise at every step and separate noise for reading and writing. The noise is part of the policy. The evaluation seed sets both the episodes and the noise stream, so the validation episodes with seed $123$ use one fixed noise stream, and the SEM over training seeds reflects variation across trained agents. Evaluation seeds $1$--$3$ give new episodes and new noise, and the ALER agents of \autoref{sec:mechanism} keep $1.00\pm0.00$ on \cInv{} and \cRSI{} there (\autoref{tab:aler-eval-ablations}). Gradients flow to every slot in proportion to its write weight. The temperatures stay fixed during training, and \autoref{tab:hyperparams_aler} lists them for each environment.

\section{\runemazes{} Detailed Description}
\label{app:rune-mazes-detailed-description}
This section specifies the observations, actions, rewards, and rune semantics of the three \runemazes{} environments. \autoref{tab:rune-configs} lists the trained configurations.

\begin{table}[htbp]
  \centering
  \small
  \caption{\textbf{Training and evaluation configurations for Rune T-Maze and Rune Multi-Corridor.} Each row is a separate training task.}
  \label{tab:rune-configs}
  \begin{tabular}{@{}l>{\raggedright\arraybackslash}p{0.6\textwidth}@{}}
    \toprule
    \textbf{Environment} & \textbf{Configuration} \\
    \midrule
    Rune T-Maze & \cInv{}, \cNoInv{}, \cSkInv{}, \cRes{}, \cRSI{} \\
    Rune Multi-Corridor & $K\in\{4,8,16\}$, no runes \\
    Rune Multi-Corridor & $K\in\{4,8,16\}$, one \cInv{} rune \\
    \bottomrule
  \end{tabular}
\end{table}

\subsection{Rune T-Maze}
\label{app:rune-t-maze}
Rune T-Maze modifies the classic T-Maze. At the start of each episode, the agent is at position $\xi=0$ of a one-dimensional corridor of normalized length $1$ and moves toward the junction at $\xi=1$. A binary cue, with $0$ for left and $1$ for right, indicates the initially correct turn. It is drawn uniformly and independently of the rune placement. The cue is visible only in the first observation and is masked to $-1$ afterwards.

\paragraph{Observations and actions.}
At each step, the agent receives the vector $o_t=[\mathit{junc},c,\mathit{dir},\mathit{rune}]$. Here $\mathit{junc}\in\{0,1\}$ equals one at the junction, $c\in\{0,1,-1\}$ is the cue, $\mathit{dir}$ is the movement direction, and $\mathit{rune}$ is the identifier of the rune just crossed. The identifiers $0,1,3,4,5$ denote \texttt{Invert}, \texttt{SkipNext}, \texttt{ResetRules}, \texttt{RepeatPrev}, and \texttt{NoOp}, and $\mathit{rune}=-1$ marks a step without a rune. The coordinate $\xi$ is not observed, and the direction stays $\mathit{dir}=1$ in the evaluated compositions. The agent has three actions, forward, turn left, and turn right. A turn has an effect only at the junction, $\xi\ge1$. Elsewhere, it costs the step penalty and does not end the episode.

\paragraph{Corridor and rune placement.}
The corridor length is $L=L_{\max}$ in fixed mode and $L\sim\mathcal{U}[L_{\min},L_{\max})$ in uniform mode, and each forward step advances the position by $1/L$. The evaluated configurations use $L=10$. Before each episode, per-type counts specify candidate runes, whose order is randomized and whose positions are drawn in $(0.02,0.98)$ with a minimum spacing of $0.1$ between candidates. Each candidate appears independently with probability $p=0.80$. A rune is consumed when crossed and cannot fire again.

\paragraph{Rune semantics.}
The environment keeps an \emph{invert} flag, a \emph{skip} flag, and the last rune that took effect. \texttt{Invert} flips the invert flag. \texttt{SkipNext} sets the skip flag. When a rune arrives while the skip flag is set, the rune is consumed without effect, the skip flag is cleared, and the last effective rune stays unchanged. \texttt{ResetRules} clears both flags and sets the direction forward. \texttt{RepeatPrev} re-applies the last effective rune, acts as \texttt{NoOp} if no rune has taken effect or the last effective rune is \texttt{RepeatPrev}, and then becomes the last effective rune. Every other rune that takes effect also becomes the last effective rune. \texttt{NoOp} changes neither flag and becomes the last effective rune. At the junction, turning right is correct if and only if $c\oplus\mathit{invert}=1$, where $c$ is the cue and $\mathit{invert}$ the invert flag.

\begin{algorithm}[t]
\caption{\textbf{ALER forward step.}}
\label{alg:aler_full}
\begin{algorithmic}[1]
\Require Observation $o_t$, memory $M_{t-1}$, recurrent state $(h_{t-1},c_{t-1})$, episode-start flag $e_t$
\If{$e_t=1$}
    \State Set $(h_{t-1},c_{t-1})$ and $M_{t-1}$ to zero
\EndIf
\State $(h_t,c_t) \gets \mathrm{LSTM}(\phi(o_t),h_{t-1},c_{t-1})$
\State $r_t \gets \texttt{Read}(h_t,M_{t-1},\tau_r)$
\State $z_t \gets \texttt{Fusion}(h_t,r_t)$
\State $M_t \gets \texttt{Write}(h_t,M_{t-1},\tau_w)$
\State $\pi_t \gets \mathrm{Actor}(\mathrm{MLP}_\pi(z_t))$, \quad $V_t \gets \mathrm{Critic}(\mathrm{MLP}_V(z_t))$
\State Sample $a_t\sim\pi_t$ in training, take $a_t=\arg\max_a\pi_t(a)$ in evaluation
\State \Return $a_t$, $V_t$, $\log\pi_t(a_t)$, $M_t$, $(h_t,c_t)$
\end{algorithmic}
\end{algorithm}

\paragraph{Rewards.}
A correct turn at the junction gives $+1$ and an incorrect turn $-1$, and both end the episode. Every other step, including forward moves and turns outside the junction, gives $-0.01$. Episodes are truncated after $T_{\max}=200$ steps.

\subsection{Rune Multi-Corridor}
\label{app:rune-multicorridor}
Rune Multi-Corridor extends Rune T-Maze to $K$ branches. The agent starts at integer position $0$ of a corridor of $L$ steps, walks to the junction at position $L$, and selects one of $K$ branches. The target branch index is drawn uniformly from $\{0,\dots,K-1\}$ independently of the rune placement and is encoded as a vector of $\lceil\log_2K\rceil$ bits, which forms the cue in the first observation and is masked to $-1$ afterwards.

\paragraph{Observations and actions.}
At each step, the agent receives $o_t=[\mathit{junc},\mathbf{b},\mathit{rune}]$ of dimension $2+\lceil\log_2K\rceil$. Here $\mathit{junc}$ equals one at the junction, $\mathbf{b}\in\{0,1,-1\}^{\lceil\log_2K\rceil}$ is the cue, and $\mathit{rune}\in\{-1,0\}$ marks an \texttt{Invert} rune at the current step. The coordinate is not observed. The agent has $1+K$ actions. Action $0$ moves forward, and action $1+k$ selects branch $k$. Branch selections have an effect only at the junction and cost the step penalty elsewhere.

\paragraph{Corridor, runes, and rewards.}
The corridor length is $L=L_{\max}$ in fixed mode and $L\sim\mathcal{U}\{L_{\min},\dots,L_{\max}\}$ in uniform mode, and episodes are truncated after $T_{\max}=2L$ steps. All evaluated configurations use fixed $L=10$. In the \cInv{} configurations, one candidate rune is placed uniformly at an interior position $\{1,\dots,L-1\}$ and appears with probability $p=0.80$. The no-rune configurations have no candidates. Crossing the rune inverts all bits of the current target, $y\leftarrow(2^{\lceil\log_2K\rceil}-1)-y$, which equals $K-1-y$ for the evaluated $K$. For example, with $K=8$ and cue $5$ (binary $101$), one \texttt{Invert} rune changes the target to $2$ (binary $010$). The agent succeeds if it selects the current target at the junction. Rewards follow Rune T-Maze.

\subsection{Rune MiniGrid Memory}
\label{app:rune-minigrid}
Rune MiniGrid Memory adds the runes of Rune T-Maze to MiniGrid Memory~\citep{chevalierboisvert2023minigridminiworldmodular}. It keeps the original action space, partial RGB observations, reward function, termination checks, and cue generation. The agent starts in the cue room facing east, and random-length maps are resampled until the requested runes fit with the required spacing. Runes lie only on corridor cells that every path to the junction crosses, so the agent always observes the cue, then the runes, then the junction.

\begin{figure}[htbp]
  \centering
  \includegraphics[width=\textwidth]{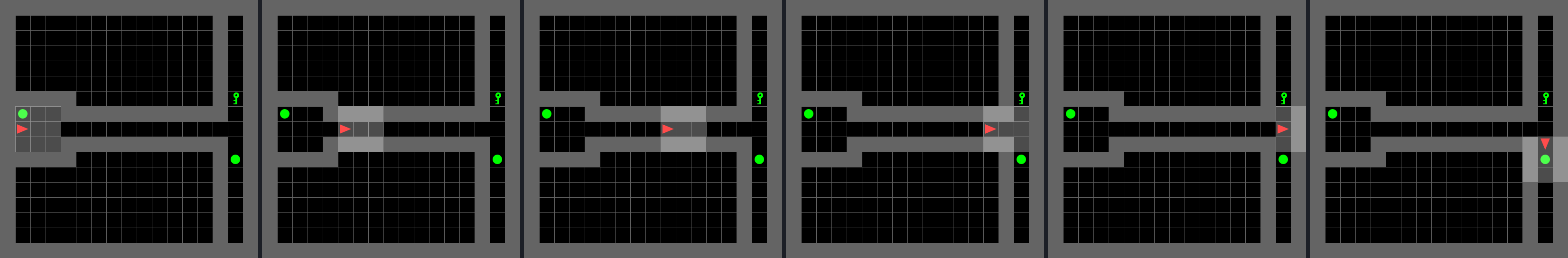}\\[2pt]
  \includegraphics[width=\textwidth]{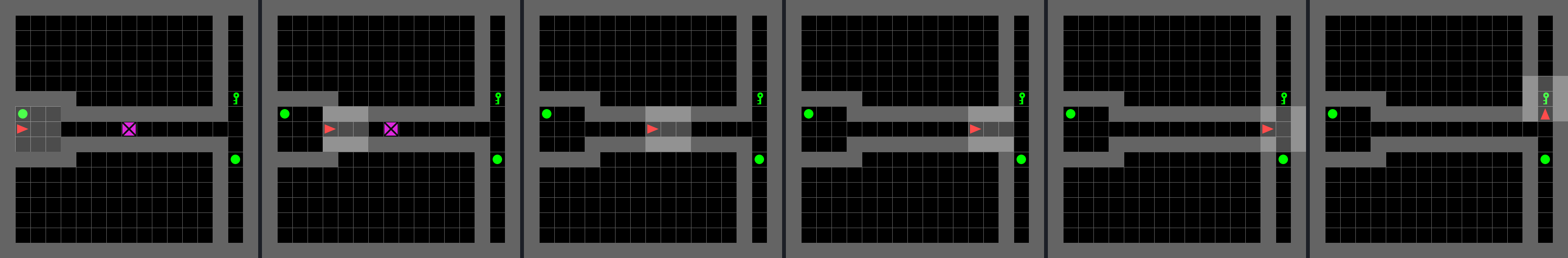}\\[2pt]
  \includegraphics[width=\textwidth]{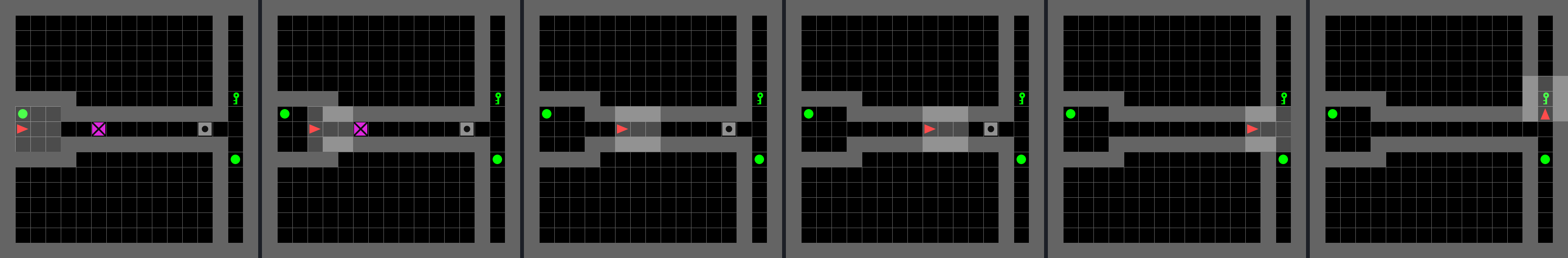}\\[2pt]
  \includegraphics[width=\textwidth]{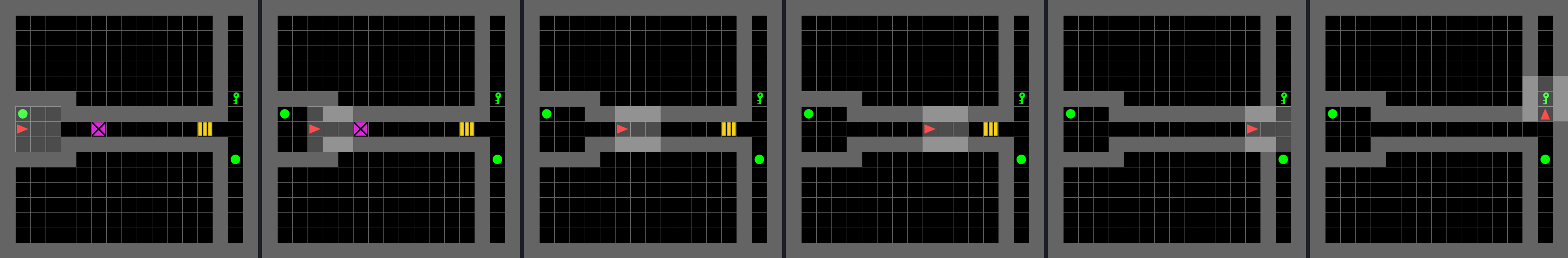}\\[2pt]
  \includegraphics[width=\textwidth]{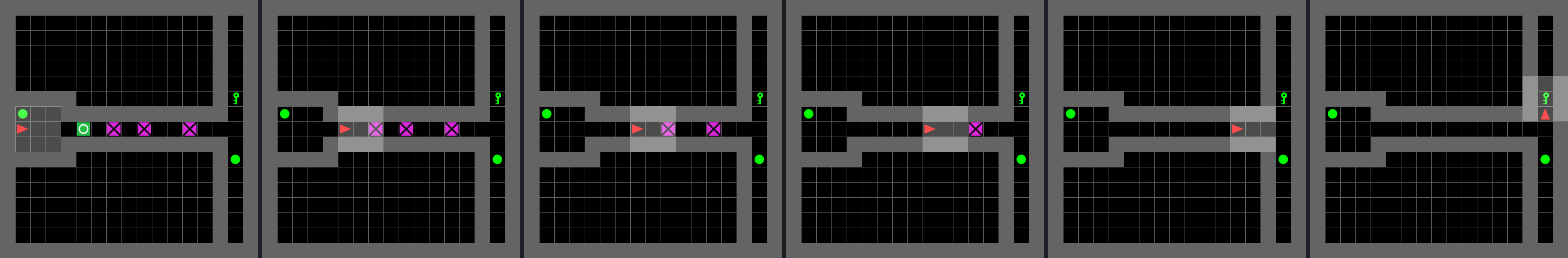}\\[2pt]
  \includegraphics[width=\textwidth]{images/rune-minigrid-repeatprev-skipnext-invert-strip.png}
  \caption{\textbf{Example Rune MiniGrid Memory rollouts.} From top to bottom: no runes, \cInv{}, \cNoInv{}, \cSkInv{}, \cRes{}, and \cRSI{}. Each strip follows one episode from the cue to the junction.}
  \label{fig:rune-minigrid-rollouts}
\end{figure}

Runes are passable floor tiles placed after the cue room and at least two cells apart. A rune can be visible before contact, fires when the agent enters its cell, is consumed, and disappears from the grid. Their semantics follow Rune T-Maze and are listed in \autoref{tab:rune-legend}. We evaluate the five Rune T-Maze compositions with fixed and random corridor lengths and train each model for $30$M environment steps with three training seeds. \autoref{fig:rune-minigrid-rollouts} shows one rollout per composition.

\begin{table}[htbp]
  \centering
  \small
  \caption{\textbf{Rune legend for Rune MiniGrid Memory.} Each rune is a floor tile of its own color.}
  \label{tab:rune-legend}
  \begin{tabular}{clp{8.2cm}}
    \toprule
    \textbf{Tile} & \textbf{Rune (color)} & \textbf{Effect} \\
    \midrule
    \raisebox{-0.25\height}{\includegraphics[height=1.6em]{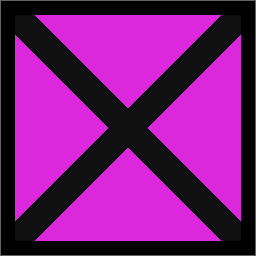}} & \texttt{Invert} (purple) & Swaps the hidden success and failure targets. The two visible objects stay in place. \\
    \raisebox{-0.25\height}{\includegraphics[height=1.6em]{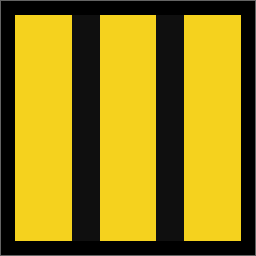}} & \texttt{SkipNext} (yellow) & Cancels the next rune the agent steps on. That rune is consumed without effect. \\
    \raisebox{-0.25\height}{\includegraphics[height=1.6em]{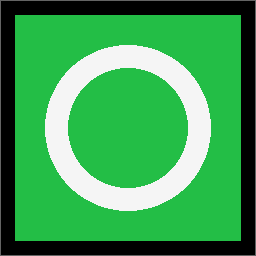}} & \texttt{ResetRules} (green) & Restores the original targets, clears the \texttt{Invert} and \texttt{SkipNext} effects, and turns the agent east. \\
    \raisebox{-0.25\height}{\includegraphics[height=1.6em]{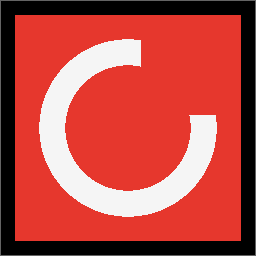}} & \texttt{RepeatPrev} (red) & Re-applies the last rune that took effect and has no effect if no rune has fired. \\
    \raisebox{-0.25\height}{\includegraphics[height=1.6em]{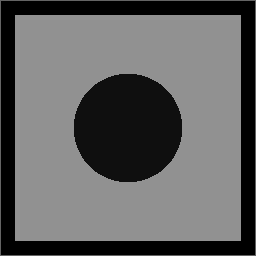}} & \texttt{NoOp} (grey) & Leaves the rules unchanged and appears as a visual distractor. \\
    \bottomrule
  \end{tabular}
\end{table}

\FloatBarrier
\section{Additional Results}
\label{app:additional-results}
This section reports Endless T-Maze with random corridor lengths, the retention-only MiniGrid Memory results cited in \autoref{sec:results}, and the learning curves of the vector environments.

\subsection{Endless T-Maze with random corridor lengths}
\label{app:endless-uniform}
\autoref{tab:endless-uniform} completes \autoref{tab:baselines} with the eight configurations in which every corridor has a random length. ALER and S5 reach $1.00$ in all eight, and PPO-LSTM reaches $1.00$ in seven and $0.98\pm0.02$ in the remaining one.

\begin{table}[htbp]
\centering
\small
\caption{\textbf{Endless T-Maze with random corridor lengths.} Each of the $n$ corridors has a length sampled uniformly from the range $l$. Success rate of the selected checkpoint on the $100$ validation episodes, mean $\pm$ SEM over five training seeds. Bold marks every value equal to the best mean in its row.}
\label{tab:endless-uniform}
\begin{adjustbox}{width=\textwidth,center}
\begin{tabular}{cccccccccc}
\toprule
\textbf{$l$} & \textbf{$n$} & \textbf{PPO-MLP} & \textbf{PPO-LSTM} & \textbf{GTrXL} & \textbf{S5} & \textbf{FFM} & \textbf{SHM} & \textbf{DNC} & \textbf{ALER} \\
\midrule
2--5 & 1 & $0.26\pm0.02$ & $\mathbf{1.00\pm0.00}$ & $0.47\pm0.08$ & $\mathbf{1.00\pm0.00}$ & $0.92\pm0.08$ & $0.55\pm0.23$ & $0.00\pm0.00$ & $\mathbf{1.00\pm0.00}$ \\
2--5 & 3 & $0.01\pm0.01$ & $\mathbf{1.00\pm0.00}$ & $0.15\pm0.09$ & $\mathbf{1.00\pm0.00}$ & $0.43\pm0.23$ & $0.04\pm0.01$ & $0.00\pm0.00$ & $\mathbf{1.00\pm0.00}$ \\
2--5 & 5 & $0.00\pm0.00$ & $\mathbf{1.00\pm0.00}$ & $0.04\pm0.01$ & $\mathbf{1.00\pm0.00}$ & $0.03\pm0.03$ & $0.04\pm0.02$ & $0.02\pm0.01$ & $\mathbf{1.00\pm0.00}$ \\
2--5 & 10 & $0.00\pm0.00$ & $\mathbf{1.00\pm0.00}$ & $0.00\pm0.00$ & $\mathbf{1.00\pm0.00}$ & $0.00\pm0.00$ & $0.00\pm0.00$ & $0.00\pm0.00$ & $\mathbf{1.00\pm0.00}$ \\
2--10 & 1 & $0.24\pm0.12$ & $\mathbf{1.00\pm0.00}$ & $0.57\pm0.02$ & $\mathbf{1.00\pm0.00}$ & $\mathbf{1.00\pm0.00}$ & $0.91\pm0.09$ & $0.00\pm0.00$ & $\mathbf{1.00\pm0.00}$ \\
2--10 & 3 & $0.00\pm0.00$ & $0.98\pm0.02$ & $0.16\pm0.03$ & $\mathbf{1.00\pm0.00}$ & $0.00\pm0.00$ & $0.06\pm0.04$ & $0.02\pm0.01$ & $\mathbf{1.00\pm0.00}$ \\
2--10 & 5 & $0.00\pm0.00$ & $\mathbf{1.00\pm0.00}$ & $0.06\pm0.01$ & $\mathbf{1.00\pm0.00}$ & $0.00\pm0.00$ & $0.04\pm0.01$ & $0.00\pm0.00$ & $\mathbf{1.00\pm0.00}$ \\
2--10 & 10 & $0.00\pm0.00$ & $\mathbf{1.00\pm0.00}$ & $0.01\pm0.01$ & $\mathbf{1.00\pm0.00}$ & $0.00\pm0.00$ & $0.00\pm0.00$ & $0.00\pm0.00$ & $\mathbf{1.00\pm0.00}$ \\
\bottomrule
\end{tabular}
\end{adjustbox}
\end{table}

\subsection{MiniGrid Memory}
\label{app:minigrid}
Before adding runes, we compare ALER and PPO-LSTM on the retention-only MiniGrid Memory tasks~\citep{chevalierboisvert2023minigridminiworldmodular} in all eight configurations, S9, S11, S13, and S17 and their random variants. ALER has the higher mean success rate in seven of the eight configurations and an average success rate of $0.84$, against $0.69$ for PPO-LSTM (\autoref{tab:minigrid-results}).

\begin{table}[htbp]
  \centering
  \small
  \caption{\textbf{MiniGrid Memory.} Success rate of the selected checkpoint on the $100$ validation episodes, mean $\pm$ SEM over three training seeds. Bold marks the higher mean in each row.}
  \label{tab:minigrid-results}
  \begin{tabular}{lcc}
    \toprule
    \textbf{Configuration} & \textbf{PPO-LSTM} & \textbf{ALER} \\
    \midrule
    S9 & $0.66\pm0.13$ & $\mathbf{0.84\pm0.22}$  \\
    S9Random & $0.83\pm0.22$ & $\mathbf{1.00\pm0.00}$  \\
    S11 & $0.55\pm0.01$ & $\mathbf{0.67\pm0.19}$  \\
    S11Random & $0.84\pm0.23$ & $\mathbf{0.95\pm0.07}$  \\
    S13 & $0.55\pm0.00$ & $\mathbf{0.83\pm0.23}$  \\
    S13Random & $0.78\pm0.16$ & $\mathbf{0.97\pm0.04}$  \\
    S17 & $\mathbf{0.55\pm0.00}$ & $0.50\pm0.05$  \\
    S17Random & $0.72\pm0.15$ & $\mathbf{0.98\pm0.03}$  \\
    \bottomrule
  \end{tabular}
\end{table}

\subsection{Learning curves}
\label{app:training-curves}
\begin{figure}[htbp]
    \centering
    \includegraphics[width=1.0\linewidth]{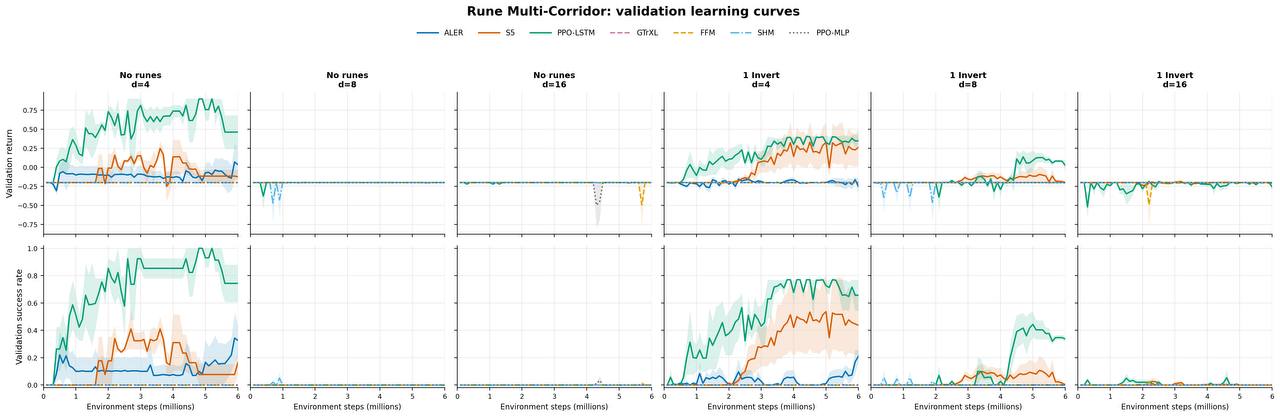}
    \caption{\textbf{Rune Multi-Corridor learning curves.} Validation return and success rate over training for ALER, S5, PPO-LSTM, GTrXL, FFM, SHM, and PPO-MLP for $K\in\{4,8,16\}$ branches, with (\texttt{rune=1}) and without (\texttt{rune=0}) an \texttt{Invert} rune.}
    \label{fig:rune-multi-training}
\end{figure}

\begin{figure}[htbp]
    \centering
    \includegraphics[width=1.0\linewidth]{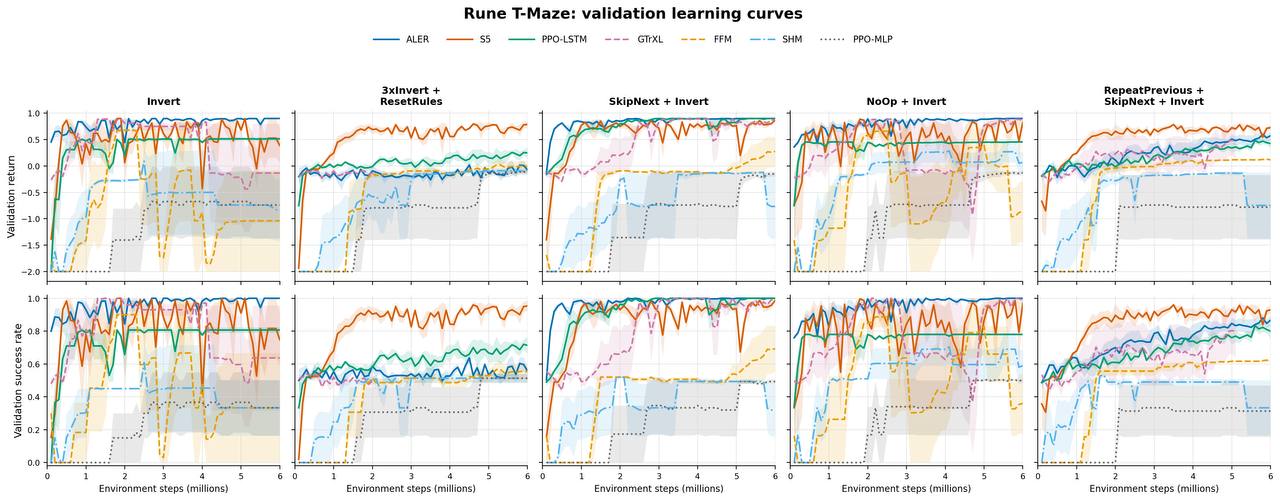}
    \caption{\textbf{Rune T-Maze learning curves.} Validation return and success rate over training for ALER, S5, PPO-LSTM, GTrXL, FFM, SHM, and PPO-MLP. Column titles give the per-type rune counts of each composition.}
    \label{fig:rune-tmaze-training}
\end{figure}

Figures~\ref{fig:rune-multi-training} and~\ref{fig:rune-tmaze-training} show the validation return and success rate over training on Rune Multi-Corridor and Rune T-Maze. Curves show the mean over training seeds, and shaded regions show the SEM.

\FloatBarrier
\section{Mechanism Analysis Details}
\label{app:ablation-details}
This section gives the protocols and full results behind \autoref{sec:mechanism}.

\subsection{Evaluation-time gate experiments}
\label{app:ablation-gate}
The gate $g_t\in[0,1]^{256}$ has one coefficient per feature. We replace it at every evaluation step by the constant $0$, $0.5$, or $1$, with frozen weights and active memory writes. These settings give the policy the recurrent state alone, an equal mixture, or the retrieved vector alone. On Endless T-Maze, we use the checkpoint trained without the coordinate input that also serves the probes (\autoref{app:ablation-probing}). Equal mixing lowers the success rate to $0.06$, $0.45$, and $0.54$ on the three tasks of \autoref{tab:aler-eval-ablations}, and the retrieved vector alone gives $0.00$, $0.26$, and $0.21$. A constant mixing ratio thus lowers success on all three tasks, and the learned feature-wise gate gives the highest success.

\begin{table}[htbp]
  \centering
  \small
  \caption{\textbf{Gate experiments on trained ALER.} Success rate, mean $\pm$ SEM over three training seeds, on evaluation seeds $1$--$3$ with $100$ episodes each. ET is Endless T-Maze with $20$ corridors of length $50$, and RT-1 and RT-2 are Rune T-Maze \cInv{} and \cRSI{}.}
  \label{tab:aler-eval-ablations}
  \setlength{\tabcolsep}{3pt}
  \begin{tabular}{@{}lccc@{}}
    \toprule
    \textbf{Gate} & \textbf{ET} & \textbf{RT-1} & \textbf{RT-2} \\
    \midrule
        Learned $g_t$ & $\mathbf{0.72\pm0.14}$ & $\mathbf{1.00\pm0.00}$ & $\mathbf{1.00\pm0.00}$ \\
        $g_t=0$ & $0.48\pm0.29$ & $0.83\pm0.02$ & $0.72\pm0.04$ \\
        $g_t=0.5$ & $0.06\pm0.06$ & $0.45\pm0.11$ & $0.54\pm0.13$ \\
        $g_t=1$ & $0.00\pm0.00$ & $0.26\pm0.13$ & $0.21\pm0.11$ \\
    \bottomrule
  \end{tabular}
\end{table}

\subsection{Linear probes}
\label{app:ablation-probing}
We fit standardized ridge classifiers ($\alpha=10$) separately for each training seed, representation, and decoded variable, on $128$ episodes per checkpoint and configuration. The agent acts greedily in these episodes, and the Gumbel noise of memory addressing stays active. Whole episodes are split $75/25$, and standardization uses only the training split.

\paragraph{Representations.}
We probe the recurrent state $h_t$ before fusion, the write candidate $x_t$, the memory before and after the write, $M_t^-$ and $M_t^+$, the retrieved vector $r_t$, the gate $g_t$, and the fused representation $z_t$. We also probe the update of the dominant write slot, $\Delta m_t=M_t^+[j_t]-M_t^-[j_t]$ with $j_t=\arg\max_j w^w_{t,j}$, which measures the change of one slot. $M_t^+$ contains all retained and newly written content.

\begin{figure}[t]
    \centering
    \includegraphics[width=\linewidth]{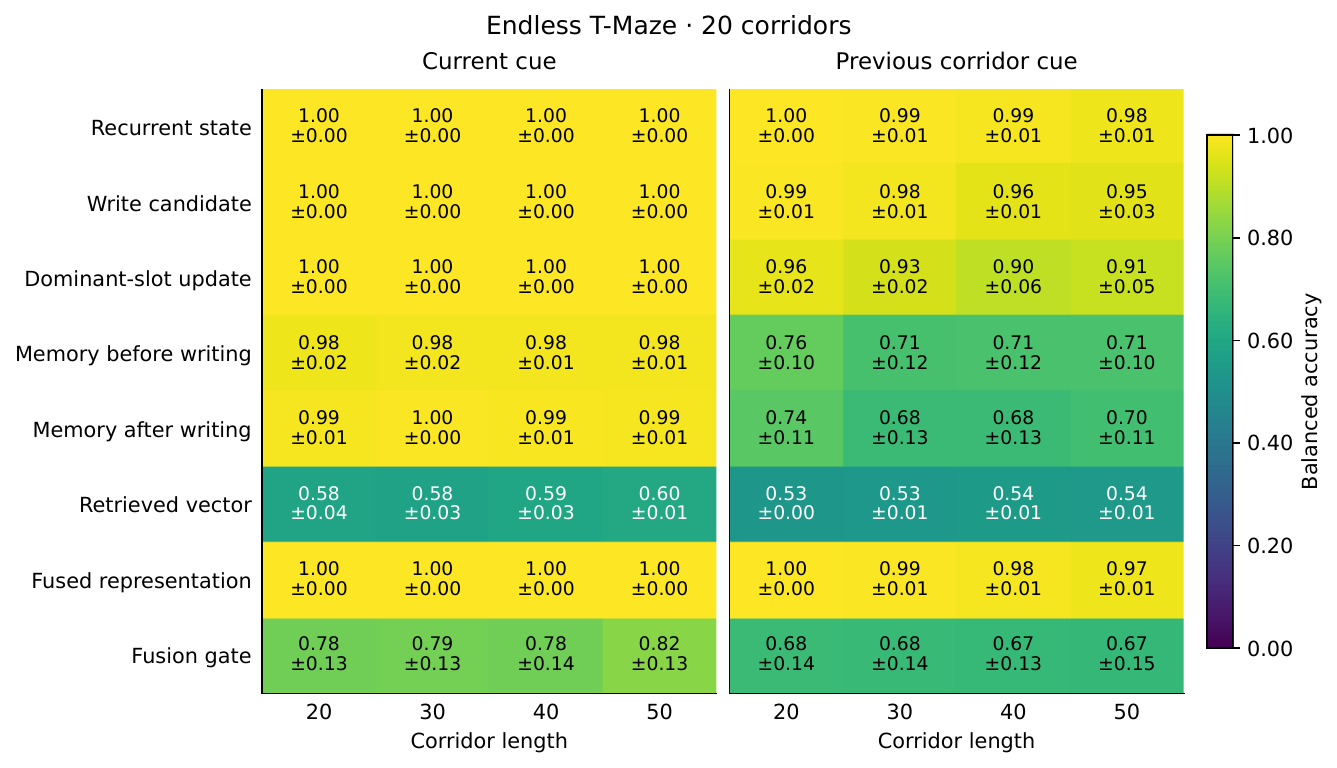}
    \caption{\textbf{Cue decoding at Endless T-Maze junctions.} Current- and previous-corridor cue decoding from eight representations with $20$ corridors of the indicated length. Each probe is fitted and tested at that length on disjoint sets of episodes. Cells report balanced accuracy, mean $\pm$ SEM over three training seeds.}
    \label{fig:aler-memory-probes}
\end{figure}

\paragraph{Endless T-Maze.}
For probing, we train ALER without the coordinate input on Endless T-Maze with $10$ corridors of lengths sampled uniformly from $2$ to $10$ and select the checkpoint on $20$ corridors of length $20$. \autoref{fig:aler-memory-probes} reports current- and previous-cue decoding at junctions for corridor lengths $20$ to $50$. Each training seed contributes one frozen checkpoint, and only the probe is refitted per length. At length $50$, the current cue is decoded with $1.00\pm0.00$ from $h_t$, $x_t$, $\Delta m_t$, and $z_t$, with $0.98\pm0.01$ from $M_t^-$, $0.99\pm0.01$ from $M_t^+$, and $0.60\pm0.01$ from $r_t$. The previous cue is decoded with $0.98\pm0.01$ from $h_t$, $0.70\pm0.11$ from $M_t^+$, and $0.54\pm0.01$ from $r_t$. The dominant read and write slots coincide on $0.08\pm0.01$ of steps, and the normalized mutual information between dominant-slot identity and either position or current cue is below $0.01$ for both operations.

\paragraph{Previous cue with the current cue held fixed.}
Consecutive cues could make the previous cue decodable through the current one. We therefore repeat junction probing on the first junction observation of corridors $2$ to $20$ and balance the four pairs of previous and current cue separately within the training and test splits, without filtering by success. We fit previous-cue probes separately for each value of the current cue and average their balanced accuracies. At length $50$, these conditional probes reach $0.99\pm0.01$ from $h_t$, $0.97\pm0.01$ from $x_t$, and $0.98\pm0.01$ from $z_t$. Shuffling the previous-cue labels within each current-cue class gives $0.48\pm0.02$, $0.51\pm0.01$, and $0.51\pm0.01$, averaged over $32$ shuffles per seed.

\begin{figure*}[t]
    \centering
    \includegraphics[width=\textwidth]{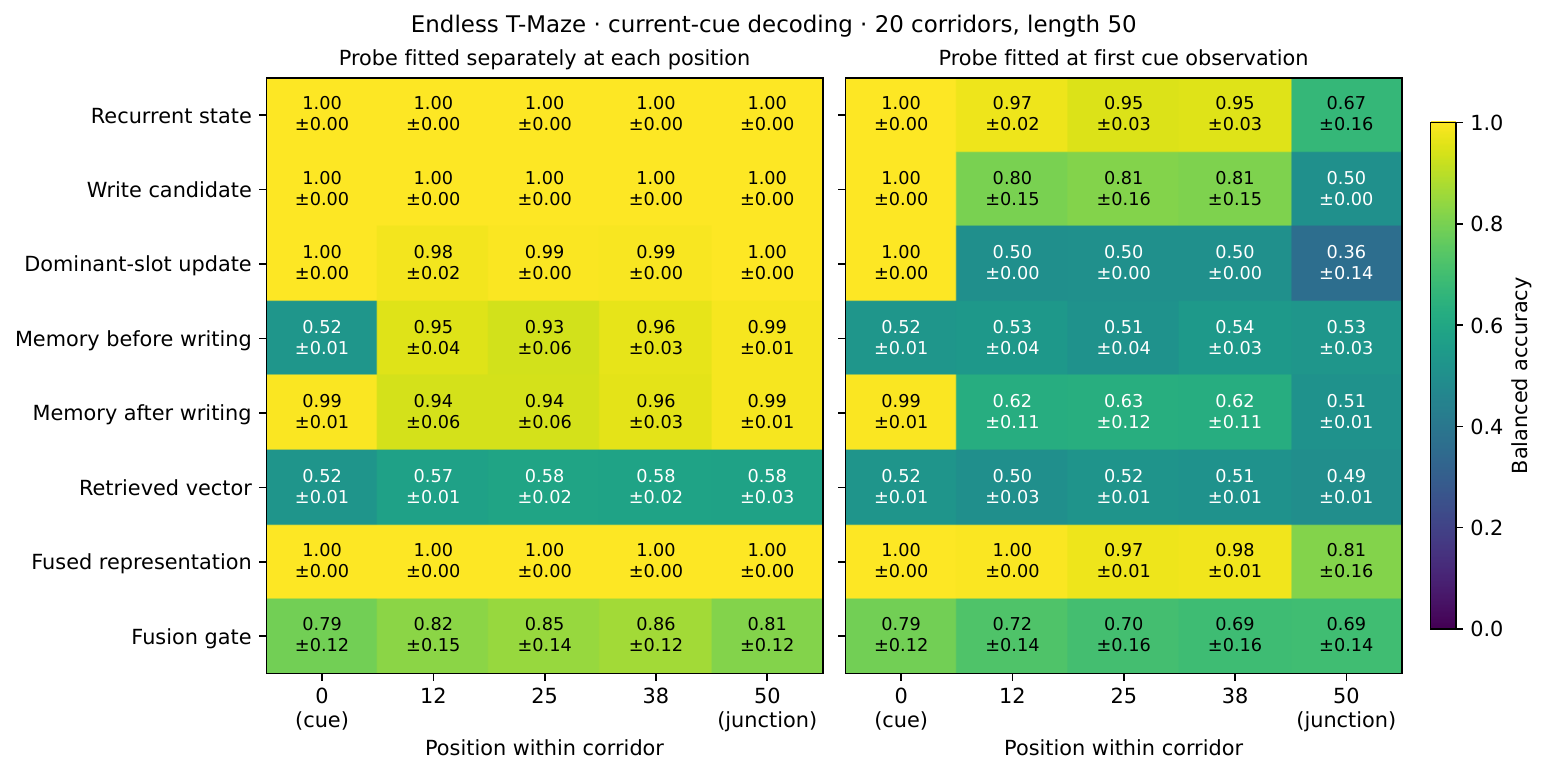}
    \caption{\textbf{Current-cue decoding within Endless T-Maze corridors.} Corridors have length $50$, with $20$ corridors per episode. Only the first observation at each sampled position is kept, and the cue is visible at position $0$ and masked at positions $12$, $25$, $38$, and $50$. Left: probes fitted at each position. Right: probes fitted at position $0$ and applied without refitting. $M_t^-$ precedes the write and $M_t^+$ includes it. Cells report balanced accuracy, mean $\pm$ SEM over three training seeds.}
    \label{fig:aler-cue-insertion}
\end{figure*}

\paragraph{Cue insertion and storage within a corridor.}
We probe positions $0$, $\mathrm{round}(L/4)$, $\mathrm{round}(L/2)$, $\mathrm{round}(3L/4)$, and $L$ at lengths $20$ to $50$ with the same checkpoints and the same balancing. Only the first observation at each position is kept, so position $0$ is the first presentation of a new cue. \autoref{fig:aler-cue-insertion} compares probes fitted at each position with probes fitted only at position $0$. At length $50$, the first cue observation raises decoding from $0.52\pm0.01$ in $M_t^-$ to $0.99\pm0.01$ in $M_t^+$, and $h_t$, $x_t$, $\Delta m_t$, and $z_t$ decode the new cue with $1.00\pm0.00$. With position-specific probes, $h_t$, $x_t$, and $z_t$ stay at $1.00\pm0.00$ at all five positions, and $M_t^+$ ranges from $0.94$ to $0.99$. Probes fitted at position $0$ reach $0.50\pm0.00$ on $x_t$ and $0.51\pm0.01$ on $M_t^+$ at the junction, so the linear code of the cue changes along the corridor, and a probe fitted at the junction reads it with $1.00\pm0.00$ and $0.99\pm0.01$.

\begin{figure}[t]
    \centering
    \includegraphics[width=\linewidth]{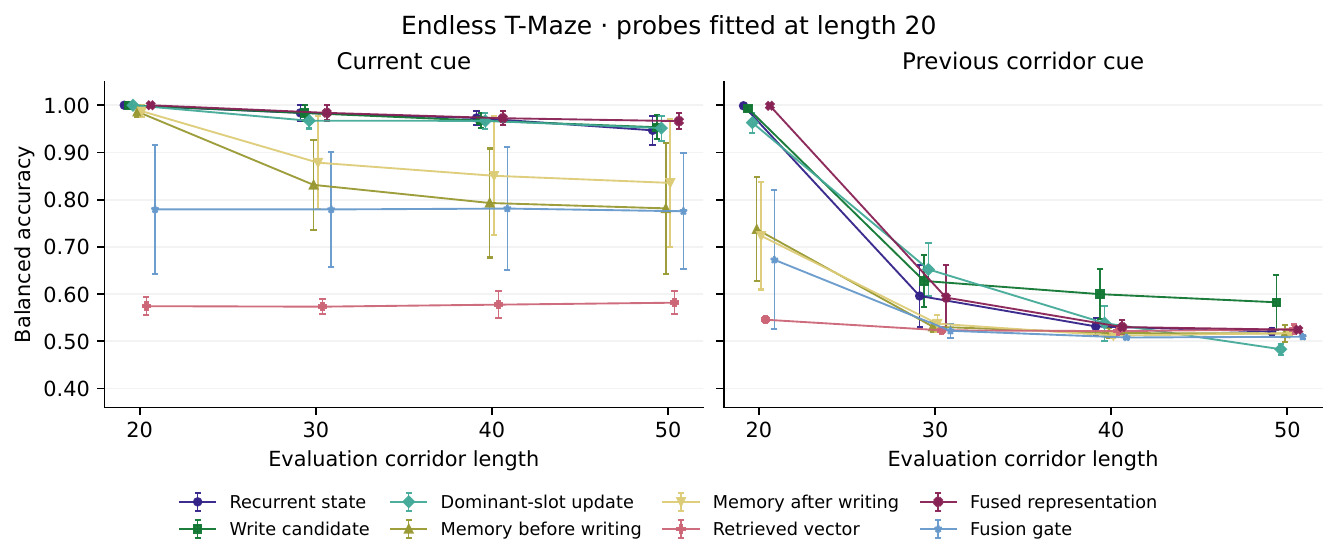}
    \caption{\textbf{Transfer of Endless T-Maze probes fitted at length 20.} Length $20$ uses held-out episodes, and longer lengths use independent sets of $128$ episodes per checkpoint. Points and error bars show the mean balanced accuracy and SEM over three training seeds, with small horizontal offsets between overlapping estimates.}
    \label{fig:aler-probe-transfer}
\end{figure}

\paragraph{Transfer across corridor lengths.}
We freeze the scaler and classifier fitted at length $20$ and apply them at lengths $30$, $40$, and $50$ (\autoref{fig:aler-probe-transfer}). At length $50$, the current cue is decoded with $0.95\pm0.03$ from $h_t$, $0.95\pm0.02$ from $x_t$, $0.97\pm0.02$ from $z_t$, $0.84\pm0.14$ from $M_t^+$, and $0.58\pm0.02$ from $r_t$. The transferred previous-cue probe on $h_t$ reaches $0.52\pm0.01$, and a probe fitted at length $50$ reaches $0.98\pm0.01$, so the current-cue code stays stable across lengths and the previous-cue code changes with the interval.

\begin{figure}[t]
    \centering
    \includegraphics[width=\linewidth]{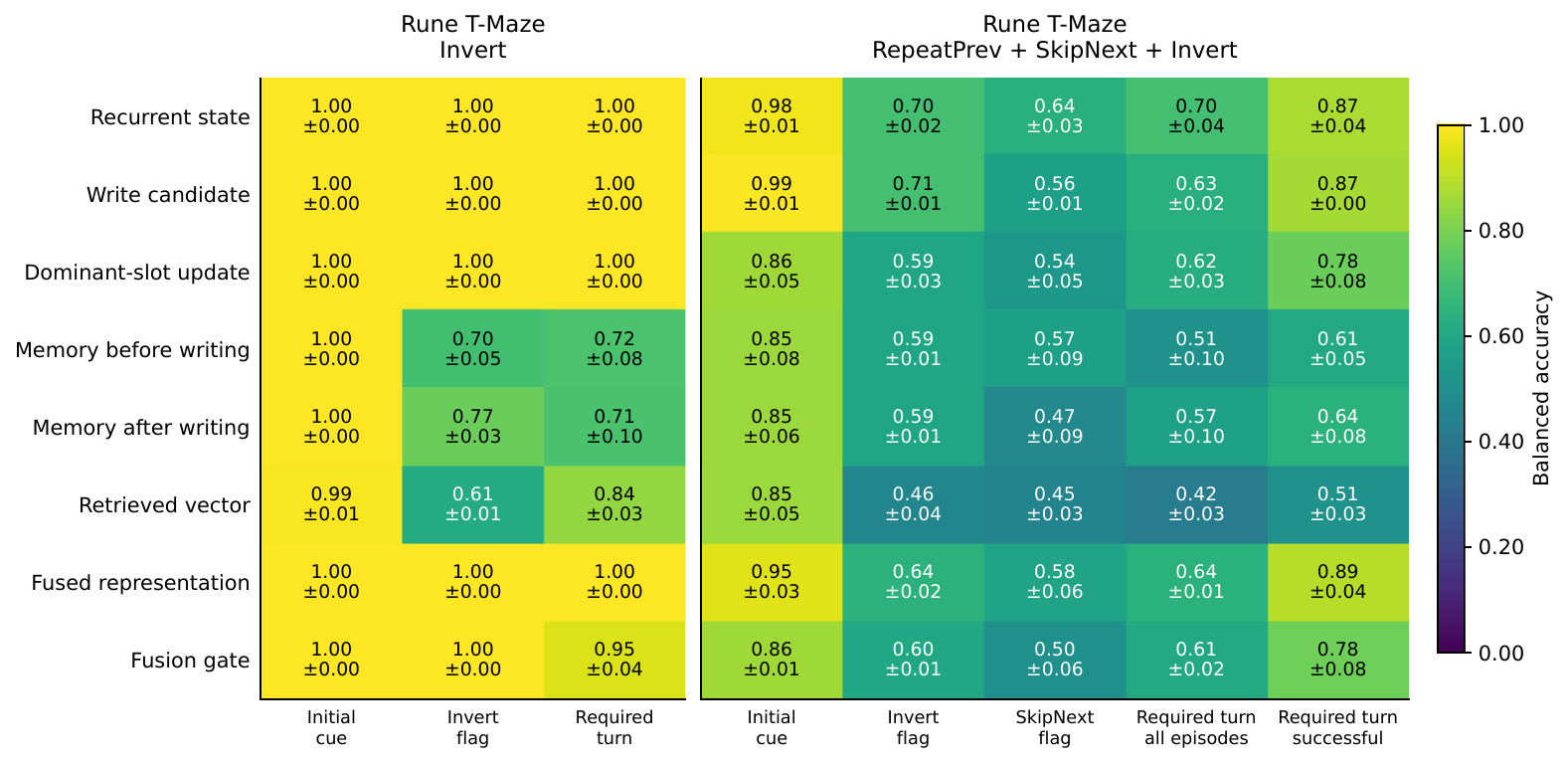}
    \caption{\textbf{Probes on Rune T-Maze.} Corridor length $10$ and rune probability $0.80$. All variables are decoded at junctions. The last column uses probes fitted and tested only on successful episodes, and the other columns use all episodes. Cells report balanced accuracy, mean $\pm$ SEM over three training seeds.}
    \label{fig:aler-rune-probes}
\end{figure}

\paragraph{Cue and rule state in Rune T-Maze.}
\autoref{fig:aler-rune-probes} reports initial-cue, rule-flag, and required-turn decoding at junctions. On \cInv{}, $h_t$, $x_t$, and $z_t$ decode all three variables with $1.00\pm0.00$, and $r_t$ decodes the required turn with $0.84\pm0.03$. On \cRSI{}, $x_t$ decodes the initial cue with $0.99\pm0.01$ and the required turn with $0.63\pm0.02$, and the update of the dominant write slot, $\Delta m_t$, gives $0.86\pm0.05$ and $0.62\pm0.03$. From $h_t$, the initial cue is decoded with $0.98\pm0.01$ and the invert and skip flags with $0.70\pm0.02$ and $0.64\pm0.03$. From $M_t^+$, the initial cue is decoded with $0.85\pm0.06$ and the two flags with $0.59\pm0.01$ and $0.47\pm0.09$. The required turn is decoded with $0.57\pm0.10$ from $M_t^+$ and $0.42\pm0.03$ from $r_t$. Probes fitted and tested only on successful episodes raise required-turn decoding from $0.70\pm0.04$ to $0.87\pm0.04$ for $h_t$, from $0.63\pm0.02$ to $0.87\pm0.00$ for $x_t$, and from $0.64\pm0.01$ to $0.89\pm0.04$ for $z_t$, and $r_t$ gives $0.51\pm0.03$. On successful episodes, the policy input $z_t$ thus carries the required turn, and the write pathway carries the initial cue on all episodes.

\subsection{Gate changes at rune observations}
\label{app:ablation-fusion}
\begin{figure}[t]
    \centering
    \includegraphics[width=\linewidth]{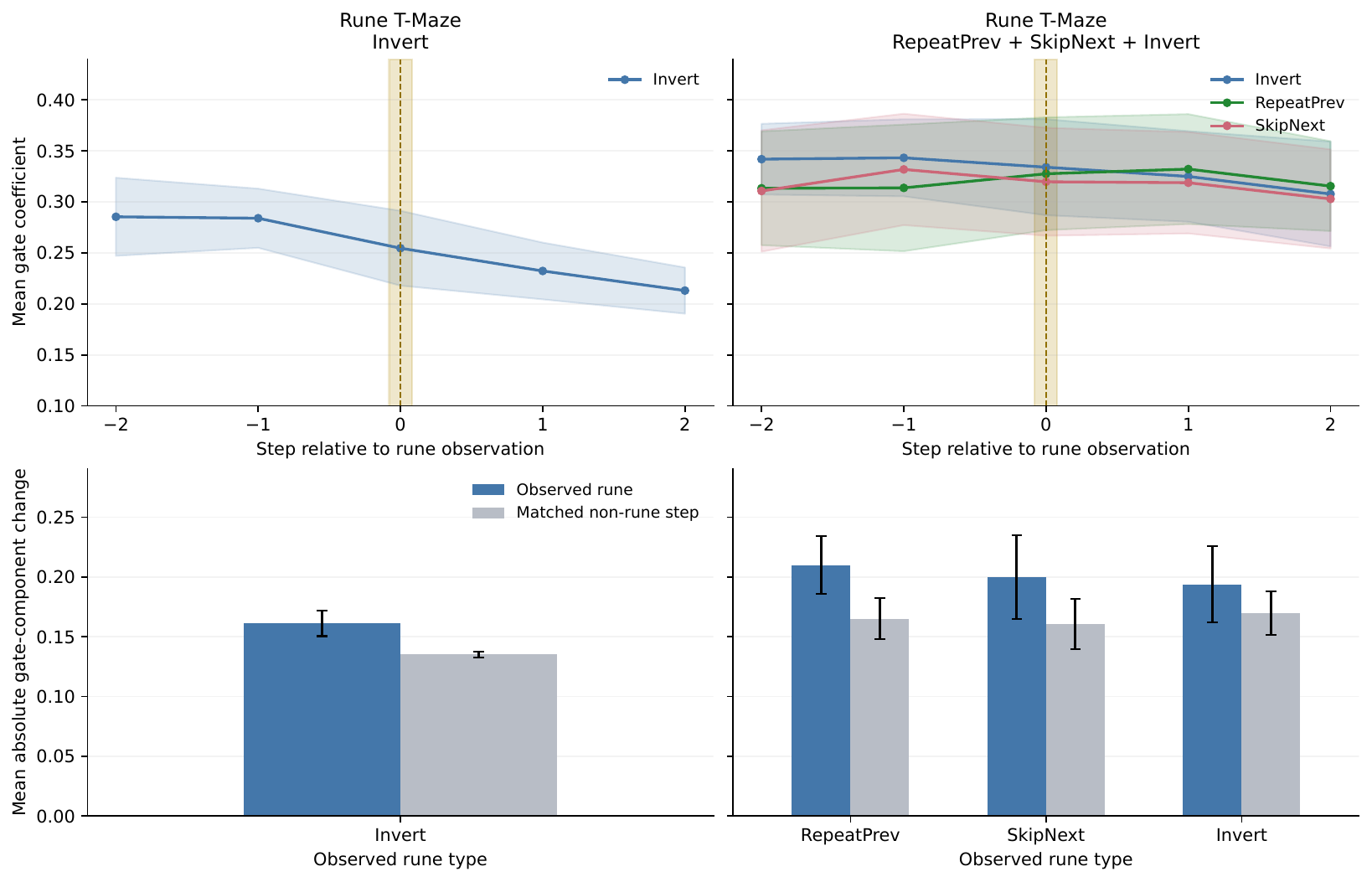}
    \caption{\textbf{Gate dynamics in Rune T-Maze.} Top: mean gate coefficient aligned to rune observations, with the rune step highlighted. Bottom: mean absolute gate-component change at rune steps and at matched non-rune steps. Shaded bands and error bars show the SEM over three training seeds.}
    \label{fig:aler-rune-fusion}
\end{figure}

We measure the change of the mean gate coefficient and the mean absolute component change $D_t=256^{-1}\sum_{k=1}^{256}|g_{t,k}-g_{t-1,k}|$. To control for the traversal phase, we match each rune step to non-rune steps with the same position, movement direction, and initial cue, excluding cue and junction steps. Position is used only for this matching and is not an input to the policy. We average within episodes, then across episodes for each seed. The excess of $D_t$ over the matched controls is $0.03\pm0.01$ at \texttt{Invert} steps of the \cInv{} composition and $0.02\pm0.02$, $0.04\pm0.02$, and $0.04\pm0.01$ at \texttt{Invert}, \texttt{SkipNext}, and \texttt{RepeatPrev} steps of \cRSI{}, and it is positive in each training seed for all four comparisons. \autoref{fig:aler-rune-fusion} shows the gate dynamics around rune observations.

\FloatBarrier
\section{Hyperparameters}
\label{app:hyper}
This section lists the selected hyperparameters of all architectures. The SHM, FFM, PPO-MLP, GTrXL, and PPO-LSTM tables describe the \runemazes{} experiments, and their Endless T-Maze configurations follow \citet{shchendrigin2026memory}. Paired values list Rune T-Maze and Rune Multi-Corridor in that order. In Rune MiniGrid Memory, the policy receives MiniGrid's egocentric $56\times56\times3$ RGB view with tile size $8$, and a three-layer CNN maps it to $128$ features before the recurrent core. ALER uses an LSTM with $128$ units, $C=5$ slots, and $\tau_r=\tau_w=0.2$. Both agents are trained by PPO with $32$ environments with rollouts of $128$ steps, minibatches of $512$, $12$ epochs, $\gamma=0.99$, $\lambda=0.98$, entropy coefficient $0.01$, a learning rate annealed from $3\cdot10^{-4}$ to $10^{-5}$, clip range $0.2$, value coefficient $0.5$, and gradient-norm clipping at $0.5$. The grid has size $17$, and validation runs every $100$k steps on $100$ episodes. PPO-LSTM uses the same CNN and PPO settings with an LSTM of $128$ units.

\begin{table}[htbp]
\centering
\begin{minipage}[t]{0.48\textwidth}
\centering
\small
\captionof{table}{\textbf{Hyperparameters for ALER on vector-based mazes.} Tuples list Endless T-Maze, Rune T-Maze, and Rune Multi-Corridor, in that order.}
\label{tab:hyperparams_aler}
\setlength{\tabcolsep}{4pt}
\begin{adjustbox}{max width=\linewidth,center}
\begin{tabular}{lc}
\toprule
\textbf{Hyperparameter} & \textbf{Value} \\
\midrule
LSTM hidden size $(d)$ & 256 \\
Number of LSTM layers & 1 \\
Memory capacity $(C)$ & 10 \\
Memory slot size & 256 \\
Key size $(d_k)$ & 256 \\
Value size & 256 \\
Read temperature $(\tau_r)$ & $0.10$ \\
Write temperature $(\tau_w)$ & $(0.20,0.10,0.20)$ \\
Rollout steps $(n_{\mathrm{steps}})$ & 128 \\
Batch size & 512 \\
PPO epochs & 12 \\
Learning rate & $(5,3,3)\!\times\!10^{-4}\!\rightarrow\!10^{-5}$ \\
Discount $(\gamma)$ & $(0.99,0.99,0.995)$ \\
GAE lambda $(\lambda)$ & 0.98 \\
Entropy coefficient & $(0.10,0.01,0.10)$ \\
Total timesteps & 6M \\
Framework & JAX \\
Parameters (vector tasks) & $\approx$1.1M \\
\bottomrule
\end{tabular}
\end{adjustbox}
\end{minipage}\hfill
\begin{minipage}[t]{0.48\textwidth}
\centering
\small
\captionof{table}{\textbf{Hyperparameters for SHM.}}
\label{tab:hyperparams_shm}
\begin{tabular}{lc}
\toprule
\textbf{Hyperparameter} & \textbf{Value} \\
\midrule
Hidden size $(h)$ & $(512,256)$ \\
Memory size $(m)$ & 64 \\
Post-processing size & 1024 \\
Learning rate & $1 \times 10^{-4}$ \\
Discount $(\gamma)$ & 0.99 \\
GAE lambda $(\lambda)$ & 1.0 \\
Entropy coefficient & 0.001 \\
Value loss coefficient & 0.5 \\
BPTT length & 1024 \\
Train batch size & 100000 \\
Minibatch size & 8192 \\
PPO epochs & 6 \\
Number of workers & 8 \\
Total timesteps & 6M \\
Framework & Ray RLlib \\
\bottomrule
\end{tabular}
\end{minipage}
\end{table}

\begin{table}[htbp]
\centering
\begin{minipage}[t]{0.48\textwidth}
\centering
\small
\captionof{table}{\textbf{Hyperparameters for FFM.}}
\label{tab:hyperparams_ffm}
\begin{tabular}{lc}
\toprule
\textbf{Hyperparameter} & \textbf{Value} \\
\midrule
Hidden size & 128 \\
Memory size $(m)$ & 128 \\
Post-processing size & 512 \\
Learning rate & $5 \times 10^{-5}$ \\
Discount $(\gamma)$ & 0.99 \\
GAE lambda $(\lambda)$ & 0.95 \\
Entropy coefficient & 0.001 \\
Value loss coefficient & 0.5 \\
BPTT length & 1024 \\
Train batch size & 100000 \\
Minibatch size & 8192 \\
PPO epochs & 6 \\
Number of workers & 8 \\
Total timesteps & 6M \\
Framework & Ray RLlib \\
\bottomrule
\end{tabular}
\end{minipage}\hfill
\begin{minipage}[t]{0.48\textwidth}
\centering
\small
\captionof{table}{\textbf{Hyperparameters for PPO-MLP.}}
\label{tab:hyperparams_mlp}
\begin{tabular}{lc}
\toprule
\textbf{Hyperparameter} & \textbf{Value} \\
\midrule
Hidden size & 128 \\
Memory & None \\
Learning rate & $1 \times 10^{-4}$ \\
Discount $(\gamma)$ & 0.99 \\
GAE lambda $(\lambda)$ & 1.0 \\
Entropy coefficient & 0.001 \\
Value loss coefficient & 0.5 \\
BPTT length & 1024 \\
Train batch size & 100000 \\
Minibatch size & 8192 \\
PPO epochs & 6 \\
Number of workers & 8 \\
Total timesteps & 6M \\
Framework & Ray RLlib \\
\bottomrule
\end{tabular}
\end{minipage}
\end{table}

\begin{table}[htbp]
\centering
\begin{minipage}[t]{0.48\textwidth}
\centering
\small
\captionof{table}{\textbf{Hyperparameters for GTrXL.}}
\label{tab:hyperparams_gtrxl}
\begin{tabular}{lc}
\toprule
\textbf{Hyperparameter} & \textbf{Value} \\
\midrule
Hidden size & 512 \\
Memory length & 100 \\
Transformer blocks & $(4,2)$ \\
Attention heads & 8 \\
Embedding dimension & 512 \\
Positional encoding & Relative \\
GTrXL gating & Enabled \\
Learning rate (initial) & $(1,2.75)\times10^{-4}$ \\
Learning rate (final) & $1 \times 10^{-5}$ \\
Discount $(\gamma)$ & $(0.995,0.99)$ \\
GAE lambda $(\lambda)$ & 0.95 \\
Gradient clipping & 0.25 \\
Entropy coeff (initial) & 0.001 \\
Entropy coeff (final) & $1 \times 10^{-6}$ \\
Value loss coefficient & 0.5 \\
PPO epochs & 3 \\
Number of environments & 16 \\
Steps per environment & 512 \\
Minibatches per epoch & 8 \\
Total timesteps & 6M \\
Framework & Custom PPO \\
\bottomrule
\end{tabular}
\end{minipage}\hfill
\begin{minipage}[t]{0.48\textwidth}
\centering
\small
\captionof{table}{\textbf{Hyperparameters for PPO-LSTM.}}
\label{tab:hyperparams_lstm}
\begin{tabular}{lc}
\toprule
\textbf{Hyperparameter} & \textbf{Value} \\
\midrule
LSTM hidden size & 512 \\
Actor network & [128, 128] \\
Critic network & [128, 128] \\
Learning rate (initial) & $3 \times 10^{-4}$ \\
Learning rate (final) & $1 \times 10^{-5}$ \\
LR schedule & Linear \\
Discount $(\gamma)$ & 0.99 \\
GAE lambda $(\lambda)$ & 0.98 \\
Gradient clipping & 0.5 \\
Entropy coefficient & 0.01 \\
Value loss coefficient & 0.5 \\
Sequence length & 128 \\
Minibatch size & 512 \\
PPO epochs & 10 \\
Number of environments & 16 \\
Steps per environment & 128 \\
Total timesteps & 6M \\
Framework & JAX \\
Parameters & $\approx$2.3M \\
\bottomrule
\end{tabular}
\end{minipage}
\end{table}

\begin{table}[htbp]
\centering
\begin{minipage}[t]{0.48\textwidth}
\centering
\small
\captionof{table}{\textbf{DNC architecture hyperparameters (POPGym implementation).}}
\label{tab:hyperparams_dnc}
\begin{tabular}{lc}
\toprule
\textbf{Hyperparameter} & \textbf{Value} \\
\midrule
Controller (LSTM) hidden size & 64 \\
Memory slots & 16 \\
Word size & 16 \\
Write heads & 1 \\
Read heads & 4 \\
Observation feature size & 64 \\
Clip value & 20.0 \\
Total timesteps & 6M \\
Framework & POPGym \\
\bottomrule
\end{tabular}
\end{minipage}\hfill
\begin{minipage}[t]{0.48\textwidth}
\centering
\small
\captionof{table}{\textbf{Hyperparameters for Resettable S5.} Tuples list Endless T-Maze, Rune T-Maze, and Rune Multi-Corridor.}
\label{tab:hyperparams_s5}
\begin{adjustbox}{max width=\linewidth,center}
\begin{tabular}{lc}
\toprule
\textbf{Hyperparameter} & \textbf{Value} \\
\midrule
S5 layers & $(2,2,4)$ \\
Model width & 256 \\
State size & $(256,128,256)$ \\
Blocks per layer & 1 \\
Activation & Full GLU \\
Normalization & None \\
Learning rate & $(5,30,30)\times10^{-5}$ \\
LR schedule & Constant \\
Discount $(\gamma)$ & 0.99 \\
GAE lambda $(\lambda)$ & 1.0 \\
Entropy coefficient & 0.0 \\
PPO clipping & 0.2 \\
Value loss coefficient & 1.0 \\
Gradient clipping & 0.5 \\
PPO epochs & 30 \\
Number of environments & 32 \\
Steps per environment & 1024 \\
Minibatch size & 8192 \\
Total timesteps & 6M \\
Framework & JAX \\
\bottomrule
\end{tabular}
\end{adjustbox}
\end{minipage}
\end{table}

\FloatBarrier
\section{Compute and LLM Usage}
\label{app:statements}
The vector-environment agents other than DNC were trained on NVIDIA H100 GPUs. ALER performs one read and one write over $C=10$ slots per step, in addition to the LSTM update. A run of $2$M environment steps takes $24.9$ minutes on average over $613$ logged runs. DNC, in the POPGym implementation, was trained on CPUs, with $462$ to $470$ minutes per run and $742$ CPU-hours in total. All GPU-based experiments reported in this work were conducted using 1 NVIDIA H100 GPU. Large language models were used for polishing and improving the clarity of the paper text.

\end{document}